\documentclass[final,5p,times,twocolumn]{elsarticle}

\usepackage{lineno,hyperref}
\modulolinenumbers[5]

\journal{Journal of Manufacturing Systems}

\usepackage{numcompress}

\RequirePackage{fix-cm}
\usepackage{graphicx}
\usepackage[caption=false]{subfig}
\usepackage{calc}
\usepackage{caption}
\usepackage{tabto}
\usepackage{booktabs}
\usepackage{listings}
\usepackage{amsmath}
\usepackage{amssymb}
\usepackage{siunitx}
\usepackage{textcomp}
\usepackage{xcolor}
\usepackage{algorithm}
\usepackage{algpseudocode}
\usepackage{svg}
\hypersetup{hidelinks} 
\usepackage{graphicx}
\usepackage[acronym]{acro}
\DeclareAcronym{asp}{
  short = ASP,
  long = Assembly Sequence Planning
}
\DeclareAcronym{dof}{
  short = DoF,
  long  = Degree of Freedom
}
\DeclareAcronym{mip}{
  short = MIP,
  long  = Mixed-Integer Program
}
\DeclareAcronym{plp}{
  short = PLP,
  long  = Production Line Planning
}

\begin{document}

\begin{frontmatter}

\title{A Unified Framework for Automated Assembly Sequence and Production Line Planning using Graph-based Optimization}


\author[A]{Christoph Hartmann\corref{corr1}\fnref{fn1}}

\address[A]{Technical University of Munich, Chair of Metal Forming and Casting, Walther-Meissner-Strasse 4, 85748 Garching, Germany}
\cortext[corr1]{Corresponding author}
\fntext[fn1]{These authors contributed equally to this work.}
\ead{christoph.hartmann@utg.de}
\author[A]{Marios Demetriades\fnref{fn1}}

\author[A]{Kevin Prüfer}
\author[A]{Zichen Zhang}
\address[B]{Faurecia Emissions Control Technologies, Germany GmbH, Biberbachstrasse 9, 86154 Augsburg, Germany}
\author[B]{Klaus Spindler}
\address[C]{Technical University of Munich, Professorship for Discrete Mathematics, Boltzmannstrasse 3, 85748 Garching b. München, Germany}
\author[C]{Stefan Weltge}



\tnotetext[1]{Abbreviations: ASP, Assembly Sequence Planning; PLP, Production Line Planning; DoF, Degree of Freedom; MIP, Mixed-Integer Program.}

\begin{abstract}

This paper presents PyCAALP (Python-based Computer-Aided Assembly Line Planning), a framework for automated Assembly Sequence Planning (ASP) and Production Line Planning (PLP), employing a graph-based approach to model components and joints within production modules. 
The framework integrates kinematic boundary conditions, such as potential part collisions, to guarantee the feasibility of automated assembly planning. 
The developed algorithm computes all feasible production sequences, integrating modules for detecting spatial relationships and formulating geometric constraints. 
The algorithm incorporates additional attributes, including handling feasibility, tolerance matching, and joint compatibility, to manage the high combinatorial complexity inherent in assembly sequence generation. 
Heuristics, such as Single-Piece Flow assembly and geometrical constraint enforcement, are utilized to further refine the solution space, facilitating more efficient planning for complex assemblies. 
The PLP stage is formulated as a Mixed-Integer Program (MIP), balancing the total times of a fixed number of manufacturing stations.
To keep the MIP tractable for complex assemblies, a deterministic path-guided reduction solves it on a subgraph assembled from complete, high-quality assembly paths, which preserves feasibility by construction and reproduces the full-graph optimum, exactly or near-exactly, at a small fraction of the directed graph and with speedups of up to three orders of magnitude.
Furthermore, the framework enables customization of engineering constraints and supports a flexible trade-off between ASP and PLP. 
The open-source nature of the framework, available at \url{https://github.com/TUM-utg/PyCAALP}, promotes further collaboration and adoption in both industrial and production research applications.

\end{abstract}


\end{frontmatter}


\section{Introduction}\label{sec:introduction}

\ac{asp} and \ac{plp} within factory planning are inherently complex, primarily because they must operate in a high-dimensional space encompassing product characteristics, production processes, and business requirements. Addressing these challenges, this paper introduces a concept to automate assembly and line planning by integrating key elements such as sequence planning, equipment selection, line configuration, and workforce allocation.

Product design has a critical influence on both assembly and disassembly operations. Consequently, frameworks like Design for Assembly (DfA) and Design for Disassembly (DfD) have become central to current research and industrial practice.
As the manufacturing industry continues shifting toward customized products, the demand for automated and adaptable process planning solutions increases.

Despite the extensive data embedded in 3D models, the potential of these resources to enhance manufacturing process planning remains vastly underutilized.
Leveraging this data from various 3D model formats could significantly optimize \ac{asp} and 
\ac{plp}, improving efficiency, adaptability, and overall effectiveness.

Another important aspect is accessibility. 
Python has become the standard language for various applications, due to its ease of use, free availability, and the variety of scientific libraries.
We believe this is the perfect combination for creating software that engineers can use and integrate with various industry workflows.  

Taking these points into consideration, the proposed Python-based framework should address these challenges by integrating both \ac{asp} and \ac{plp} problems into a unified approach.
It incorporates engineering, geometric, and manufacturing constraints to enhance realism and applicability.
A \ac{mip}-based formulation is used under the assumption of a fixed number of workstations. 
The framework is designed to be data-efficient, user-friendly, and adaptable to real production environments.
Additionally, it introduces a novel method for modeling geometric constraints by extracting individual degree-of-freedom matrices from a 3D model through joint analysis.

\section{State of the art}\label{sec:stateoftheart}
Existing research on \ac{asp} and \ac{plp} focuses on developing computational methods that address their inherent complexity and provide efficient problem-solving algorithms.

\subsection{Assembly planning}

Various approaches have been developed to solve the \ac{asp} problem, i.e., determining the optimal assembly sequence given specific objectives or constraints.
Most rely on a directed graph representation of the assembly, typically derived from a solid model.
The graph nodes, representing components, may have a contact, blocking, or free relation with the other nodes in the graph. 
Early contributions utilized \ac{dof} matrices to enforce geometric constraints and capture the node relations for any pair of the assembly components and their surfaces. 
The matrix elements represent the available \ac{dof} in both positive and negative directions for translation and rotation.
Furthermore, these contributions employed stability and accessibility constraints to secure parts and manage tool access restrictions \citep{laperrire_al_1996_GAPP_a_generative_assembly_process_planner}.
The same constraints remain relevant in more recent methods.
For instance, Xia et al  \cite{xia_lu_lu_al_2024_Semantic_knowledge_driven_A_GASeq_A_dynamic_graph_learning_approach_for_assembly_sequence_optimization} proposed a dynamic graph learning algorithm that integrates heuristic information into sequence optimization processes, improving performance and adaptability in complex assembly cases.

Other recent approaches have utilized deep learning methods for more complex assemblies (i.e., those comprising more than 1000 entities) to develop knowledge-enhanced graph embedding models \citep{jing_zhou_zhang_al_2024_XMKR_Explainable_manufacturing_knowledge_recommendation_for_collaborative_design_with_graph_embedding_learning},    \citep{Shi_Xiaolin_Tian_al_2022_Knowledge_graph_based_assembly_knowledge_towards_complex_product_assembly_process}. 
In addition, Shi et al. \cite{Shi_Tian_Ma_Wu_Gu_2024_A_knowledge_graph_based_structured_representation_of_Assembly_process_planning_combined_with_Deep_Learning} applied named entity recognition to sequentially classify text data and output the type of various entities. 
Despite successfully embedding complex assemblies, these efforts do not directly solve an \ac{asp} problem.
At the same time, many deep learning-based approaches require large, labeled datasets.
Since these data are task-specific and company-dependent, their generalization is difficult.
Thus, these methods are practically limited in highly customized environments.

Finally, multiple methods exist for finding the optimal sequence given an assembly graph. 
They include shortest path algorithms \citep{Laperrière_ElMaraghy_1994_GAPP2_Assembly_sequences_planning_for_Simultaneous_Engineering_Applications}, greedy algorithms with feedback weights
\citep{Zhu_Xu_Wang_Yang_Fan_2023_Graph_based_assembly_sequence_planning_algorithm_with_feedback_weights}, and fitness-score based techniques on engineering constraints \citep{xia_lu_lu_al_2024_Semantic_knowledge_driven_A_GASeq_A_dynamic_graph_learning_approach_for_assembly_sequence_optimization}.

\subsection{Line planning and time balancing}

The \ac{plp} problem typically presents as an optimization problem that aims to distribute production process steps across various stages, while satisfying operational constraints.
The most common approach is a \ac{mip} model, formulated based on constraints such as cycle time limits, precedence relations, and station capacities.
The objective is to maximize the process output or the total cost for a given set of resources.

Beyond that, a wave of metaheuristic approaches complements the traditional exact methods.
Genetic algorithms, simulated annealing, ant colony optimization, and particle swarm optimization have gained prominence due to their flexibility and effectiveness in handling complex and large-scale problems \citep{line-bal-review}.
In addition, \cite{li.zhi.2024} implemented chance-constrained stochastic models combined with a branch-and-bound and remember algorithm, which ensures solution feasibility under probabilistic constraints.
Although not guaranteed to produce optimal solutions, most of these methods are simple, fast, and robust, often yielding high-quality results, making them suitable for modern variable and complex manufacturing environments \citep{Ahmad.2024}.

\section{Problem Formulation}
\label{sec:problem}

While significant advances have been made in both \ac{asp} and \ac{plp}, relatively few works attempt to integrate these aspects into a unified workflow.
Our work aims to tackle this challenge and integrate both problems.

Formally, the problem transforms the physical assembly definition into a search space optimization defined by three components:

\begin{enumerate}
    \item \textbf{Assembly Graph:} The part geometry is modeled as a connected undirected graph $G_{part} = (N, J)$, where $N$ represents the set of components (nodes) and $J$ represents the set of joints (edges).
    
    \item \textbf{Weighted Directed Graph:} The solution space is mapped to a layered directed graph $D = (V, E, W)$. 
    Here, $V$ represents the assembly states (cutsets), and $E$ the directed edges, corresponding to a joint operation between cutsets. 
    The set of weights $W$ assigns a weight $w_e$ to each operation $e \in E$, derived from the engineering constraints.
    
    \item \textbf{Objective Function:} The optimization goal is to minimize the global cost $\mathcal{J}$, defined as a weighted sum of the soft engineering constraints' cost ($\mathcal{C}_{eng}$) and the time balancing cost ($\mathcal{C}_{time}$).
    The two parts correspond to the \ac{asp} and the \ac{plp}, respectively:
    \begin{equation}
        \mathcal{J} = (1 - \lambda) \cdot \mathcal{C}_{eng} + \lambda \cdot \mathcal{C}_{time}
        \label{eq:obj_func}
    \end{equation}
    where $\lambda \in [0, 1]$ is the user-defined influence factor controlling the trade-off between assembly sequence quality (engineering constraints) and assembly line efficiency (station time).
\end{enumerate}

Therefore, given an input graph $G_{part}$ and a fixed number of manufacturing stations, the objective is to find a path in $D$ that minimizes $\mathcal{J}$ subject to engineering and geometrical constraints.
The engineering constraints (i.e., part fragility, manufacturing tolerances, and technological changes) are incorporated into the edge weights $w_o$, while the geometrical constraints define the feasible operations $o$ when connecting a joint.

The assembly sequence and station time minimization problem of Eq.~\ref{eq:obj_func} is solved using a \ac{mip}, given the weighted directed graph $D$ as input.
In addition to the user-defined influence factor $\lambda$, the user can also control the relative weights of specific engineering constraints within $w_o$.

The ultimate goal is to find solutions fast and efficiently. 
For typical assemblies, \ac{asp} and \ac{plp} problems should be computed within minutes.
For more complex assemblies, the framework applies graph-reduction techniques to reduce the size of $D$, yielding an optimal or near-optimal solution in reasonable computational time.
The joint count alone does not determine tractability, as the difficulty of the resulting \ac{mip} depends on the specific graph structure, especially for higher $\lambda$ values.
In our test cases, the reduction becomes beneficial for the 17-joint Assembly 2.

\section{Approach}\label{sec:approach}
\begin{figure}[h]
    \centering
    \includegraphics[width=1.0\linewidth]{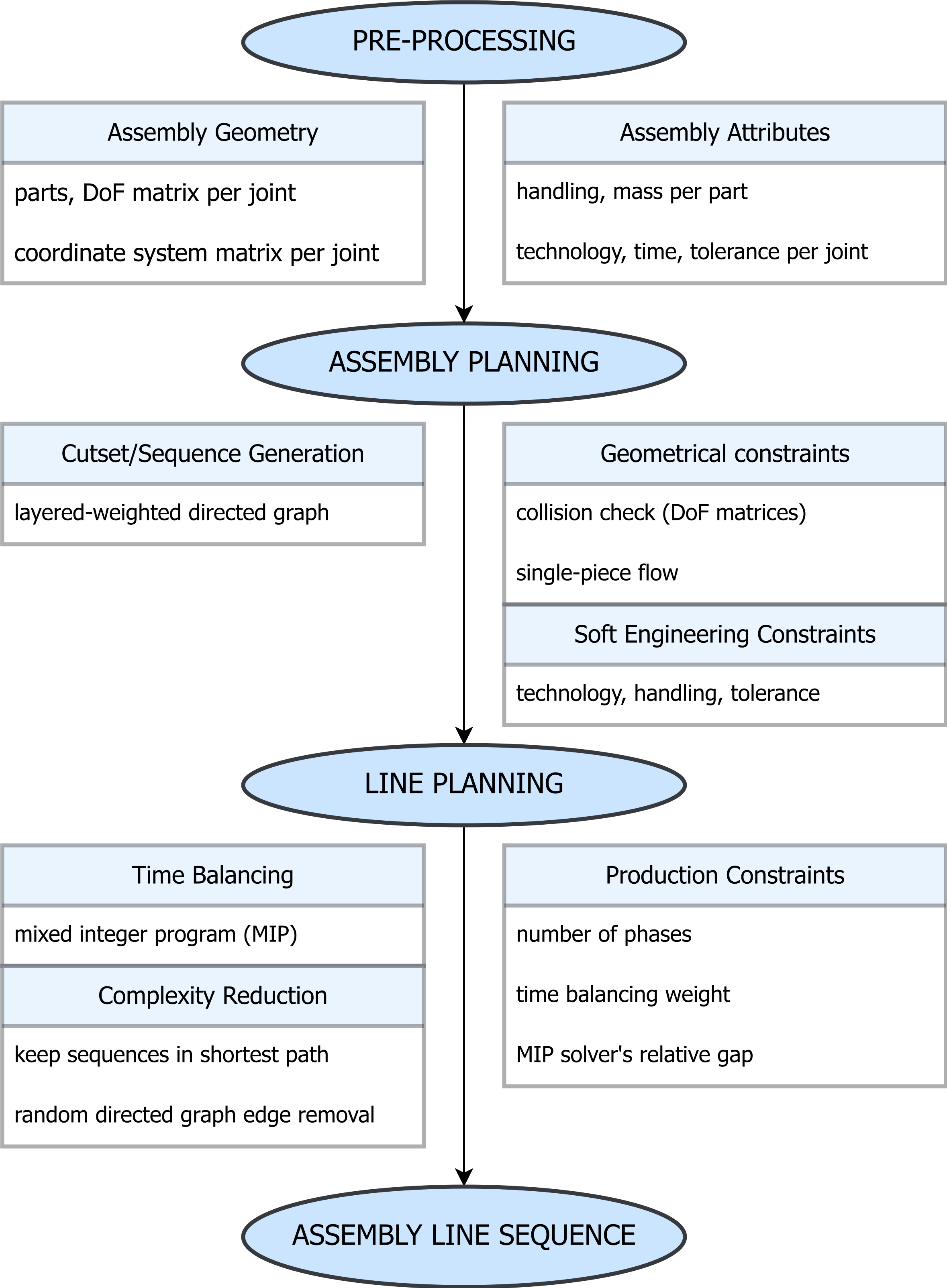}
    \caption{Workflow of the proposed unified framework. The process flows from CAD data pre-processing (extracting joints and \ac{dof} matrices) to Assembly Planning (generating the weighted directed graph with soft engineering and geometric constraints) and finally to Line Planning (solving the \ac{mip} for time balancing).}
    \label{fig:workflow}
\end{figure}

Our approach tries to solve both \ac{asp} and \ac{plp} simultaneously.
It consists of three steps: pre-processing, assembly planning, and line planning.
Figure~\ref{fig:workflow} illustrates the complete workflow, including functionalities and attributes/constraints per step.
Ultimately, the best assembly path is calculated considering all constraints, with the option to prioritize either assembly or line planning.

\subsection{Pre-processing}
In the first step, we utilize CAD models to automatically extract spatial information for use in the subsequent assembly planning step.
Specifically, joints, \ac{dof} matrices, and coordinate systems are extracted for each joint.
Furthermore, the assembly attributes (handling, mass per part, technology, time, and tolerance per joint) are assumed to be known.

\subsection{Assembly planning}
In the assembly planning step, all the possible assembly cutsets (an assembly connectivity state) and sequences are calculated while enforcing geometrical and engineering constraints.
The geometrical constraints are a combination of a heuristic and a collision check with \ac{dof} matrices. 
The user-defined technology, handling, and tolerance weights are used to construct a weighted directed graph, controlling the influence of these attributes.
The output is a layered, weighted directed graph consisting of all the geometrically feasible sequences.  

\subsection{Line planning}
In the final step, a time balancing problem, formulated as a \ac{mip}, solves the \ac{plp}.
The user-defined production constraints determine the problem formulation, solution accuracy, and the influence of assembly and line planning in the solution. 
Furthermore, for complex cases, an optional operation uses a deterministic path-guided edge reduction, solving the \ac{mip} on a subgraph comprising a percentage of the weighted directed graph.
Finally, optimal assembly sequence and manufacturing stages planning can be extracted from the \ac{mip}'s solution.
The user can run the workflow multiple times for various engineering and production constraints.
In this case, the pre-processing stage should be executed only once.
Thus, once the model data is available, only the assembly planning and time balancing stages need to be repeated.

\section{Methods}\label{sec:methods}
This section provides a more in-depth coverage of the steps mentioned in Section \ref{sec:approach} and illustrated in Fig~\ref{fig:workflow}.
More precisely, the procedures, algorithms, and design of the Python-based framework are discussed.

\subsection{Assembly part connection}

\begin{figure}[h]
    \centering
    \includegraphics[width=1.0\linewidth]{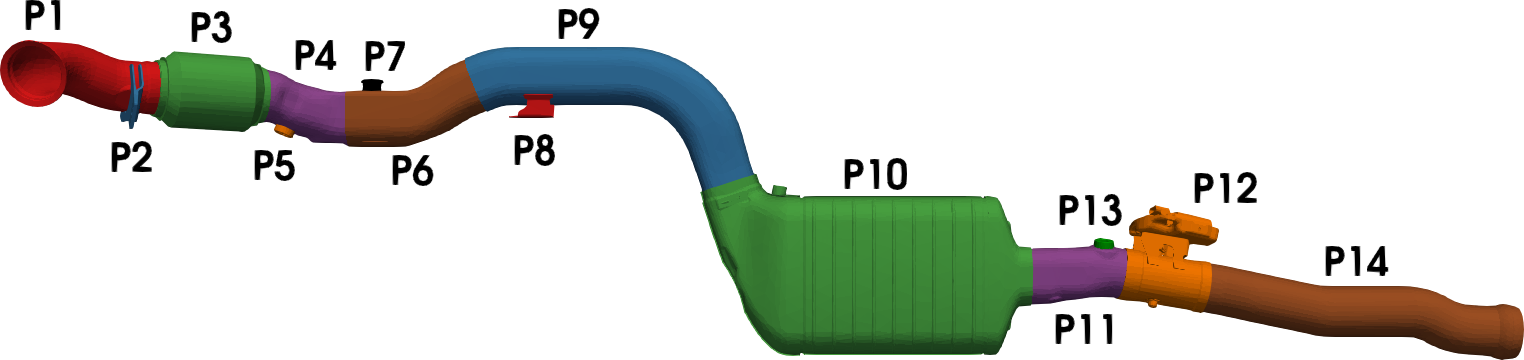}
    \caption{Assembly 1 geometry and part labels.}
    \label{fig:ass_1_labels}
\end{figure}

\begin{figure}[h]
    \centering
    \includegraphics[width=0.48\textwidth]{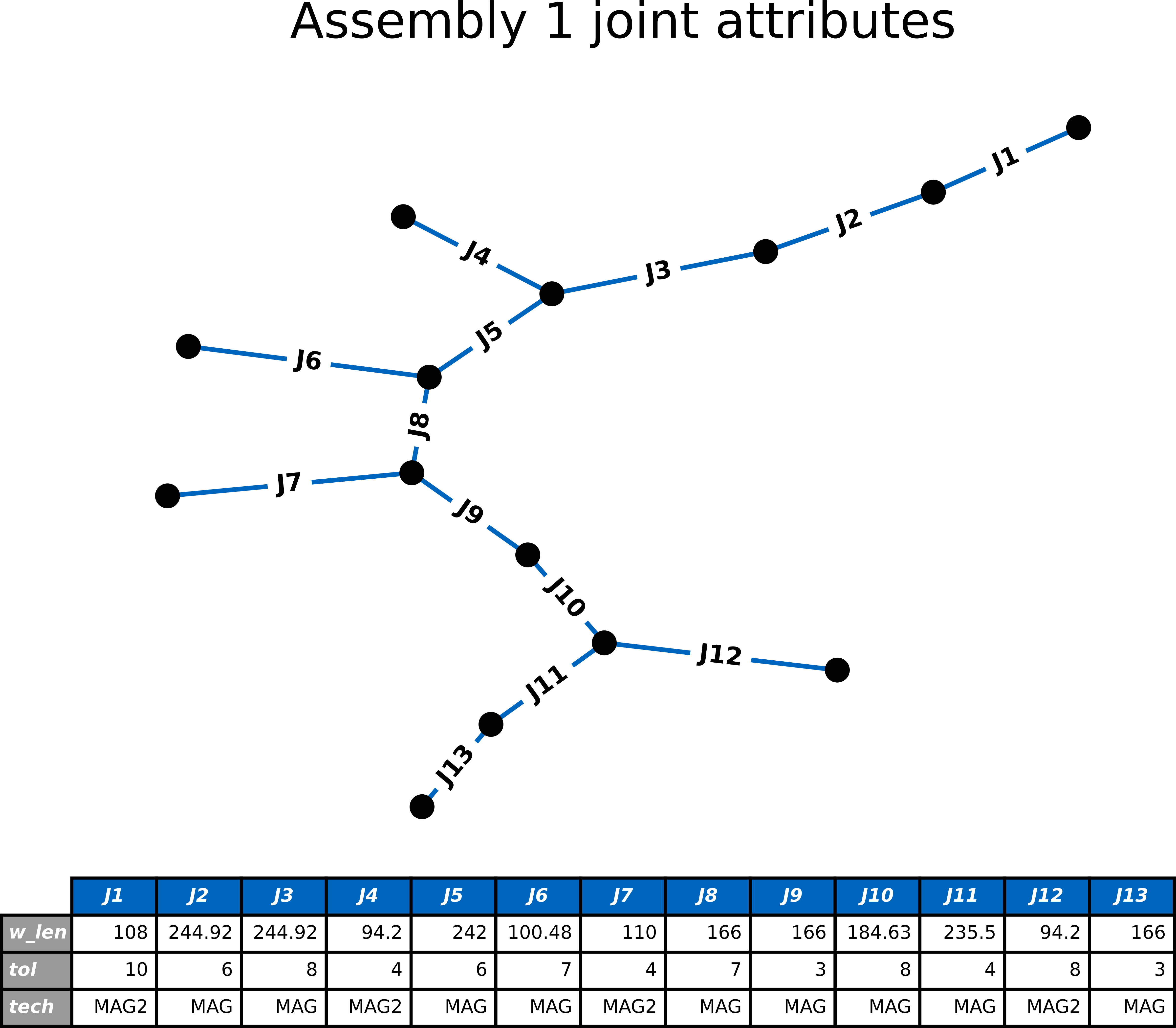}
    \caption{Graph representation of the joints (edges) for Assembly 1. Edge labels indicate the welding length [mm] (used as operation time in our model), tolerance value (integer 1-10, where higher values indicate stricter tolerances), and the joining technology type ("MAG" vs "MAG2")}
    \label{fig:ass_1_edges}
\end{figure}

\begin{figure}[h]
    \includegraphics[width=0.48\textwidth]{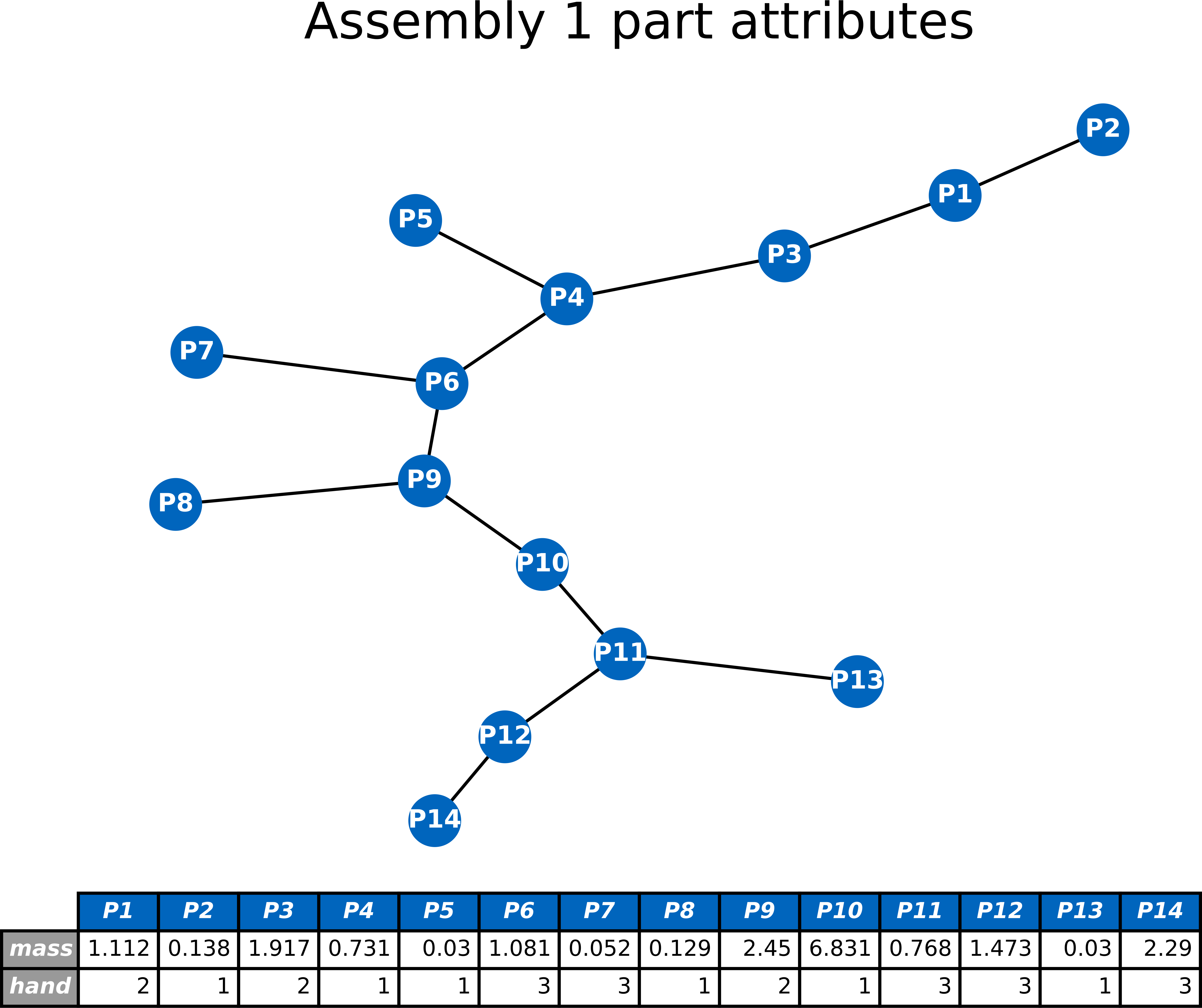}
    \caption{Graph representation of the parts (nodes) for Assembly 1. Node labels indicate component mass [kg] and the handling complexity level (normalized integer 1-3).}
    \label{fig:ass_1_nodes}
\end{figure}

In the first part of pre-processing, the assembly part connections are represented as an undirected graph, where nodes correspond to the assembly parts and edges correspond to the joints.
The part information includes part names, weights, and handling instructions.
Similarly, the joint information includes parts of the joint, time, and tolerance.
Figures \ref{fig:ass_1_nodes} and \ref{fig:ass_1_edges} show the graph representation of one of the two industrial assemblies (see Fig. \ref{fig:ass_1_labels}) used for testing, along with the absolute values of the attributes, i.e., non-normalized, for joints or parts.

In practice, the part data are provided in a JSON file, and the graph is represented using \texttt{NetworkX} library \citep{Aric_al_netx}. 
All the attributes are then normalized to control the contribution to the overall weight of the next step's weighted directed graph.

\subsection{Degree of freedom matrices}
The second part of the pre-processing deals with the \ac{dof} matrices. 
They are optionally extracted from the CAD files for use in enforcing geometric constraints, i.e., collision checking.
This part consists of two operations: spatial relationship detection and geometrical constraint construction.

First, we export the CAD data in STL format, which uses a triangle mesh representation, storing the surfaces of the CAD models as a set of triangles. 

The graphics libraries for the CAD part include \texttt{PyVista} (used for basic mesh opeartions), \citep{sullivan2019pyvista}, \texttt{Flexible Collision Library (FCL)} \citep{python-fcl}, \texttt{MeshLib} \citep{meshlib}, and \texttt{trimesh} \citep{trimesh}.

\subsubsection{Spatial relationship detection}
To obtain a reasonable assembly sequence, the spatial relationships among the CAD models should be taken into account. 
There are three types of spatial relationships defined in this framework: contact relationships, blocking relationships, and free relationships. 
An appropriate data structure to store the spatial relationship information between each pair of CAD models is a graph. 
We denote this graph as a relational graph.

The three kinds of spatial relationships are defined as follows:
\begin{enumerate}
    \item Two CAD models have a contact relationship if they physically interact.
    \item Two CAD models have a blocking relationship if they do not have physical contact, but they will have collisions when translating one of them in some orientations.
    \item Two CAD models have a free relationship if their relationship is neither contact nor blocking.  
\end{enumerate}

In practice, we implement the relational graph as a matrix. 
We denote this matrix as a relational matrix. 
Since the spatial relationships are mutual, which means the spatial relationship between A and B is the same as that between B and A, the relational graph is undirected, and the relational matrix is symmetric.

Afterwards, we use \texttt{FCL} to detect contact relationships and \texttt{PyVista} to detect blocking relationships.
\texttt{FCL} is employed to detect whether two given CAD models have collisions.
The two CAD models will have a contact relationship if they have collisions.
If the relationship is not a contact relationship, we use an approach based on the raytracing method from \texttt{PyVista} to detect whether the relationship is blocking.
This approach will cast rays in predefined orientations starting from each vertex of one CAD model and detect whether these rays intersect with the other CAD model.
If there is at least one ray from any of the vertices that intersects with the other CAD model, the two CAD models will have a blocking relationship.

\subsubsection{Geometric constraint construction}

Furthermore, additional geometric constraints are crucial to ensure that the assembly sequence is reasonable.
These geometric constraints are in the format of a \ac{dof} matrix, which gives more detailed spatial information among the CAD models.

A \ac{dof} matrix, as in Eq.~\eqref{eq:dof_matrix_example}, is a three-by-four matrix containing twelve elements representing the corresponding translational and rotational \ac{dof}. 
$T$ and $R$ are translational \ac{dof} and rotational \ac{dof} matrices, respectively. 
There are actually three translational \ac{dof} and three rotational \ac{dof} in a 3D space.
In this framework, we split each \ac{dof} into two, based on orientations. 
For instance, the translational \ac{dof} matrix in x orientation is split into one translational \ac{dof} in positive x orientation and one translational \ac{dof} in negative x orientation.

\begin{equation}
    M = \begin{pmatrix}
     T_{+x} & T_{-x} & R_{+x} & R_{-x} \\
     T_{+y} & T_{-y} & R_{+y} & R_{-y} \\
     T_{+z} & T_{-z} & R_{+z} & R_{-z}
    \end{pmatrix}
    \label{eq:dof_matrix_example}
\end{equation}

Note that these \ac{dof} matrices are only constructed for CAD model pairs with a contact or blocking relationship.
To obtain these \ac{dof}, corresponding geometric movements are applied to a CAD model from a given pair, and the intersection volume between the moved CAD model and the other CAD model from this pair is computed using \texttt{MeshLib}.
If the volume exceeds the tolernace, the corresponding \ac{dof} for the moved CAD model will be zero.
For example, we apply a geometric translational movement in the positive x orientation to a CAD model.
If the intersection volume between it and the other CAD model is not zero, then the translational \ac{dof} in the positive x orientation for the moved CAD model is zero.
The translational displacements are not fixed constants but are generated adaptively for each part pair: for every axis, the projected characteristic length of the moved part is scaled by a range factor of $2.0$ and divided into five equal steps, yielding test distances of $0.4$--$2.0\times$ the characteristic length, applied in both the positive and negative directions.
The rotational tests use six pre-defined angles ($15^{\circ}$, $45^{\circ}$, $75^{\circ}$, $90^{\circ}$, $120^{\circ}$, $150^{\circ}$), applied in both rotational senses.
A \ac{dof} entry is set to zero whenever \emph{any} tested displacement produces an intersection volume above a tolerance of $0.1$~\si{\cubic\milli\metre}.
Additional test magnitudes can therefore only zero further entries, never create spurious freedoms, so the procedure cannot admit a geometrically infeasible operation.
Because the translational range is derived from the geometry of each pair rather than fixed a priori, the test scales automatically with part size and requires no per-assembly tuning.
A small example of \ac{dof} matrix extraction is shown in \ref{app:dof_case}.

Furthermore, the \ac{dof} matrices and a unique coordinate system per joint are provided in JSON format to serve as geometrical constraints during the assembly planning process.   

\subsection{Cutset/sequence generation}

After gathering all important assembly information, the generation of the cutset/sequence layered digraph follows.
In this step, the cutsets are represented by weighted directed graph nodes, where the number of layers equals the number of assembly joints + 1, i.e., one joint addition per layer.
Each edge corresponds to a joint addition, and each node to a unique cutset state. 
A cutset state corresponds to a distinct structural connectivity of the partially assembled product.
The weights of the graph's edges are calculated using the user-defined weighted constants ($\mu$) for technology, handling, and tolerance (see Eq. \ref{eq:edge_weight}). 
These constants give control over the assembly planning. 

Let $\mu_{tech}, \mu_{hand}, \mu_{tol} \in [0, 1]$ be the user-defined weights such that $\mu_{tech} + \mu_{hand} + \mu_{tol} = 1$ and $l_e \in [1, L] $ be the directed graph's layer of edge $e$.
The calculation for the weight $w_e$ of an edge $e$, assuming that all attribute weights are normalized, is given by:
\begin{equation}
    w_e = \frac{\mu_{tech} \cdot w_{tech}(e) + \mu_{hand} \cdot w_{hand}(e) + \mu_{tol} \cdot w_{tol}(e)}{l_e}
    \label{eq:edge_weight}
\end{equation}

We assume that higher handling and tolerance values indicate greater fragility or tighter tolerances.
Furthermore, dividing by the current layer number, $l_e$, introduces a penalty for risks occurring in earlier stages.
This formulation enforces these risks, i.e., technology changes, higher handling requirements, and tighter tolerances, to appear in later stages of the layered graph.
In the absence of normalization by $l_e$, the positional influence of the operations is lost, as the \ac{mip} optimizes only for the total accumulated weight.
Given that the attributes per joint are constant, the resulting weights for every possible sequence remain identical.

The directed graph is generated using a two-step fused method.
Initially, using the connected assembly graph, all the possible cutsets are obtained by removing one edge at a time.
For each step, i.e., edge removal, one layer is created in the directed graph, and all of its nodes consist of assembly graphs with an equal number of edges. 
The final node of the directed graph is the disconnected assembly graph. 
Thus, the directed graph is generated starting from its last layer, the fully connected part.  

Equation \ref{eq:max-edges} shows the maximum possible number of edges $|E|$ of the directed graph, for a given number of assembly joints ($E$).
The solution space of a directed graph created by a relatively big assembly increases dramatically.
For example, a directed graph representing an assembly with 17 joints might have more than 1.1 million edges.
A slightly more complex assembly, consisting of 20 joints, might have ten million edges.
Trying to solve the \ac{mip} of the \ac{plp} stage may result in an unreasonable solving time.
Therefore, geometric constraint methods are used to reduce the solution space.
Engineering constraints or heuristics are applied when creating new edges in the directed graph.
First, with the removal of an edge, the resulting assembly graph should consist of a single subassembly, meaning that the actual manufacturing process operates with only one assembly, thereby enforcing Single-Piece Flow.

\begin{equation}
   |E|_{max}=\sum_{j=0}^{J-1}(J-j)\binom{J}{j}= J \cdot 2^{J-1}  
\label{eq:max-edges}
\end{equation}

\subsection{Collision check}
In addition, using the \ac{dof} matrices, a collision scenario is checked, excluding infeasible edge removals and enforcing the geometrical constraints. 
First, this check is optional and available only for assemblies with generated \ac{dof} matrices from the CAD data. 
We use this functionality later for assembly 2.
For assembly 1, this step is skipped.

\noindent \textbf{Initialization:} Before the directed graph generation, for every joint $j$, a local coordinate system is extracted from the CAD data.
A $4 \times 4$ homogeneous transformation matrix $T_j$ is constructed by combining the translation matrix $D_j$ and the rotation matrices $R_{x,j}, R_{y,j}, R_{z,j}$:
\begin{equation}
    T_j = D_j \cdot R_{x,j} \cdot R_{y,j} \cdot R_{z,j}
\end{equation}
where $D_j$ and $R_{\cdot,j}$ are defined as homogeneous matrices:
\begin{equation}
    D_j = \begin{bmatrix} I_3 & \mathbf{t}_j \\ \mathbf{0}^T & 1 \end{bmatrix}, \quad
    R_{\cdot,j} = \begin{bmatrix} \mathcal{R}_{\cdot,j} & \mathbf{0} \\ \mathbf{0}^T & 1 \end{bmatrix}
\end{equation}
Here, $I_3$ is the $3 \times 3$ identity matrix, $\mathbf{t}_j$ is the translation vector to the origin $O_j$, and $\mathcal{R}_{\cdot,j}$ represents the standard $3 \times 3$ rotation matrix around the respective principal axis \citep{craig2005intro_rob}.
These matrices serve as the basis for calculating relative spatial relationships in the subsequent steps:

\begin{enumerate}
    \item \textbf{Neighbor identification:} Upon incorporating a new joint $j_{new}$ and its corresponding component $n_{new}$ into the current subassembly cutset $v$, the algorithm identifies the set of incident joints $j_{inc}$ connected to $n_{new}$. 
    A neighbor joint $j_{neigh} \in j_{inc}$ is selected for geometric verification if and only if its connecting part $n_{neigh}$ is already present in the active subassembly ($n_{neigh} \in v$). 
    This condition ensures that collision checks are performed only against components that physically exist in the current partial assembly state.
    \item \textbf{Reference frame setup:} Set the transformation matrix of the new joint, $T_{ref}$, as the reference frame. 
    Calculate the relative transformation matrix, $T_{rel}$, required to map the neighbor's coordinate system ($T_{neigh}$) into this reference frame:
    \begin{equation}
        T_{rel} = T_{ref}^{-1} \cdot T_{neigh}
    \end{equation}
    This transformation ($T_{rel}$) represents the spatial difference between the existing subassembly component and the newly added joint.
    
    \item \textbf{\ac{dof} transformation:} Transform the neighbor's \ac{dof} vectors, $v_{neigh} \in \mathbb{R}^3$, into the reference coordinates. 
    Since DoF vectors represent directions, they are treated as homogeneous vectors with a zero fourth component, $v_{hom} = [v_{neigh}^T, 0]^T$:
    \begin{equation}
        v'_{neigh} = \left| \left( T_{rel} \cdot v_{hom} \right)_{1:3} \right|
    \end{equation}
    The operation returns the element-wise absolute values of the translational components (the first two columns of the freedom matrix), filtering out values below a numerical tolerance threshold ($\epsilon \approx 10^{-15}$) to address floating-point noise.
        
    \item \textbf{Collision check:} Perform the collision check by comparing available translational movements.
    The new joint is deemed feasible if and only if it allows at least the same degrees of freedom as the neighbor requires in the transformed orientation.
    This is evaluated using the $L_0$ norm (count of non-zero elements):
    \begin{equation}
    || v_{ref} ||_0 \ge || v'_{neigh} ||_0 \quad \forall v \in {T{+}, T-}
    \end{equation}
    where $v_{ref}$ corresponds to the reference joint's DoF vector and $v'_{neigh}$ is the transformed vector from the previous step.
\end{enumerate}

Finally, all the edges in the cutset-directed graph have a single calculated weight, which will be used in the next steps for the reduction of directed graph complexity and for the \ac{mip}. 
A complete algorithm is presented in \ref{app:algo}, detailing all the steps required to create the assembly-weighted directed graph.

\subsection{Deterministic path-guided edge reduction}
\label{subsec:det_red}


The geometrical constraints remove infeasible operations, but the weighted directed graph can still be too large for the \ac{mip}.
For instance, the directed graph of Assembly 2 retains $134{,}216$ edges, and solving the full \ac{mip} to optimality requires between a few minutes and roughly $4.7$~h, depending on the time-balancing factor $\lambda$ (Table~\ref{tab:speedup}).

Rather than pruning $D$ directly, we solve the \ac{mip} on a \emph{subgraph} $D'=(V',E')\subseteq D$ built from a set $\mathcal{P}$ of complete, high-quality paths, where
\begin{equation}
    E' = \bigcup_{p\,\in\,\mathcal{P}} \operatorname{edges}(p),
    \qquad
    V' = \bigcup_{p\,\in\,\mathcal{P}} \operatorname{nodes}(p).
    \label{eq:subgraph}
\end{equation}
The construction is fully deterministic and reproducible.
Since every $p\in\mathcal{P}$ is a complete $v_{start}\!\rightarrow\!v_{end}$ path, $D'$ always admits a feasible flow (Eqs.~\eqref{eq:x.1}--\eqref{eq:x.3}) and, because $D'\subseteq D$, its optimum is a feasible solution of the full problem and hence an upper bound on the full-graph optimum.
The reduction can therefore never return an infeasible or invalid assembly sequence, at any budget.
An unstructured reduction that simply discards low-weight edges offers no such assurance: it can sever every $v_{start}\!\rightarrow\!v_{end}$ path and leave the \ac{mip} infeasible.
What the construction does \emph{not} guarantee is that the optimal path is among those enumerated.
The design question is therefore not whether to keep paths but which paths to enumerate, so that a small $\mathcal{P}$ already contains the globally optimal sequence, or a close approximation; how closely the reduced optimum tracks the full one is assessed empirically in Section~\ref{subsec:scalability}.

We evaluate six enumeration strategies, all of them built on Yen's algorithm for the $k$ shortest simple paths~\cite{yen1971kshortest}, as implemented in \texttt{NetworkX}'s \texttt{shortest\_simple\_paths}, or on a deterministic
penalized variant of it.
\begin{itemize}
    \item \textbf{Engineering weights (\textit{edge\_w})}: The engineering weights, i.e., the directed graph's weights $w_e$ of Eq.~\eqref{eq:edge_weight}, are used to calculate $k$ shortest paths.
    This targets assembly-sequence quality but ignores line balancing.
    
    \item \textbf{Balance weight (\textit{bal\_w})}: Edges that cross an ideal phase boundary (a multiple of the phase width $T/P$) are favored, prioritizing paths whose cumulative time splits cleanly into equal-load phases.
    The phase width is the ideal station time: $\frac{T_O}{P}$.
    
    \item   \textbf{Combined}: To increase the coverage of solution quality across $\lambda$ values, the two previous methods (\textit{edge\_w} and \textit{bal\_w}) are combined.
    
    \item   \textbf{Blended}: A single per-edge weight parameterized by the \ac{mip} factor $\lambda$,
        \begin{equation}
           w^{\text{bl}}_e(\lambda) \;=\;
           (1-\lambda)\,\widehat{w}_e \;+\; \lambda\,\widehat{m}_e,
           \label{eq:blended}
        \end{equation}
    where $\widehat{w}_e$ is the min--max-normalized engineering weight and $\widehat{m}_e$ is a normalized, path-additive phase misalignment proxy. 
    Each ideal boundary $b_k=k\,T/P$ is crossed by exactly one edge of any path (cumulative time is monotone), and that edge is charged the best attainable cut error $\min(b_k-t_u,\,t_v-b_k)/(T/P)$, i.e.\ how far the nearer endpoint, i.e., the only place a phase can actually be cut, lies from the ideal boundary.
    Enumerating by Eq.~\eqref{eq:blended} targets the $\lambda$-specific compromise path directly, instead of unioning the two corner enumerations and relying on one of them being right.
    The misalignment term is a heuristic surrogate for the true objective (which uses the maximum phase time $\alpha$, not a path sum).
    
    \item   \textbf{Blended union (\textit{bl-union})}: Eq.~\eqref{eq:blended} enumerated over a grid of $\lambda$ values to cover the edge cases $\lambda\!\in\!\{0,1\}$, with the resulting edge sets unioned.
  
    \item   \textbf{Diverse - penalty method}: A deterministic penalized re-routing where, after each shortest path is found, a fixed penalty is added to its edges, forcing the next shortest path onto new edges. 
    In this method, each iteration is a single linear-time shortest-path computation, which is cheaper than Yen's algorithm for a high number of paths.
    
\end{itemize}

Two of these strategies are individually the strongest but have complementary weaknesses. 
\texttt{bl-union} discovers high-quality paths, but its edge coverage \emph{saturates}: beyond a point, additional paths
reuse edges already in $E'$ and the subgraph stops growing.
\texttt{Diverse} grows without bound, but its penalty ordering can skip high-quality paths. 
We therefore combine them into an \texttt{adaptive} strategy: enumerate with \texttt{bl-union} until its edge growth stalls (a fixed number of consecutive rounds add no new edge), then switch to \texttt{diverse} penalized rerouting to keep expanding $E'$ until the size budget is met.
The adaptive strategy is deterministic throughout.

\textbf{Stop criterion:} Enumeration stops when $|E'|$ reaches a user-defined fraction of $|E|$, a target subgraph percentage, rather than a fixed path count $k$. 
A percentage budget is preferable because the mapping from $k$ to the number of retained edges is strongly instance- and method-dependent, whereas a subgraph fraction is directly comparable across assemblies and controls the size of the reduced \ac{mip}.
As shown in Section~\ref{subsec:scalability}, a small budget already captures the optimum, so the percentage should be set conservatively low rather than large.

\subsection{Time balancing}
The final part of the procedure addresses the \ac{plp} problem for a given number of manufacturing stations.
This problem is tackled with the use of a \ac{mip} and the cost function in Eq. \ref{eq:obj_func}.
The formulated \ac{mip}'s objective function is presented in Eq.~\eqref{eq:obj} and the constraints from Eq.~\eqref{eq:x.1} to \eqref{eq:alpha.1}. 
The current problem attempts to assign each operation, i.e., joint, to the correct phase, i.e., manufacturing station, considering the assembly directed graph's edge weights and the maximum phase time.

\begingroup
\setlength{\abovedisplayskip}{3pt}
\setlength{\belowdisplayskip}{3pt}
\footnotesize
\begin{align}
\label{eq:obj} \text{min} \quad & (1 - \lambda) \sum_{e \in E} (w_e x_e) + \lambda c \alpha \\
\label{eq:x.1} \text{s.t.} \quad & \sum_{e \in \text{out\_edges}(v_{start})} x_e = 1 \quad &&  \\
\label{eq:x.2} & \sum_{e \in \text{in\_edges}(v_{end})} x_e = 1 \quad &&  \\
\label{eq:x.3} & \sum_{e \in \text{in\_edges}(v)} x_e = \sum_{e \in \text{out\_edges}(v)} x_e \quad && \forall v \in V \setminus \{v_{start}, v_{end}\} \quad  \\
\label{eq:y.1} & \sum_{p=0}^{P-1} y_{l,p} = 1 \quad && \forall l \in \{0, \dots, L-1\} \quad \\
\label{eq:y.2} & y_{l,0} \leq y_{l-1,0} \quad && \forall l \in \{1, \dots, L-1\} \quad  \\
\label{eq:y.3} 
& y_{l,p} \leq y_{l-1,p-1} + y_{l-1,p}
&& \forall l
, p \in \{1, \dots, P-1\} \\
\label{eq:z.1} & \sum_{p=0}^{P-1} z_{o,p} = 1
&& \forall o \in O \quad  \\
\label{eq:xyz.1} 
& z_{o(e),p} \geq x_e + y_{l(e),p} - 1 
&& \forall e \in E, \ p \in \{0, \dots, P-1\} \\
\label{eq:alpha.1} 
& \alpha \geq \sum_{o \in O} z_{o,p} \cdot \text{t}_o \quad 
&& \forall p \in \{0, \dots, P-1\}
\end{align}
\endgroup

\noindent
We define the notation as follows:
$w_e$ is the assembly-directed graph weight of edge $e$.
$\lambda \in [0,1]$ is a user-defined time-balancing coefficient which controls the contribution of \ac{asp} and \ac{plp}.   
$E$ is the set of edges in the directed graph. 
$V$ is the set of nodes (cutsets) in the directed graph.
$O$ is the set of physical operations (joints). 
$P$ is the number of phases (stations).
$L$ is the number of layers. 
$l(e)$ denotes the layer index to which edge $e$ belongs.
$o(e)$ denotes the operation corresponding to edge $e$.
$t_o$ is the operation time required for operation $o$.
$x_e \in \{0,1\}$ indicates whether edge $e$ is used.
$y_{l,p} \in \{0,1\}$ assigns directed graph's layer $l$ to phase $p$.
$z_{o,p} \in \{0,1\}$ assigns operation $o$ to phase $p$. 
$\alpha$ denotes the maximum time of any phase.
$c$ is an equal contribution factor that is pre-calculated to enforce equal contribution of assembly and line planning parts, i.e., in the objective value. 
The calculation of $c$ is given by:

\begin{equation}
    c
    = \frac{\displaystyle \sum_{e\in p^*}w_{e}}
    {\displaystyle \frac{1}{P}\sum_{o\in O}t_o}
    \label{eq:c}
\end{equation}
where $p^*$ denotes the engineering-weight shortest path of the directed graph (\textit{edge\_w}), and the numerator sums over its edges.
The factor is computed from the optimal standalone solutions of both \ac{asp} (shortest path) and \ac{plp} (average phase time).
Finally, the two terms are balanced and the user can control their tradeoff with $\lambda$.

The constraints can be described as follows:
The first three constraints deal with the $x$ variable.
Equation~\eqref{eq:x.1} requires the sum of all outgoing edges of the start node $v_{start}$ to be one.
Similarly, Eq.~\eqref{eq:x.2} enforces the sum of all ingoing edges of the end node $v_{end}$ to be one.
Equation~\eqref{eq:x.3} ensures the flow conservation for all the other nodes, i.e., the sum of ingoing and outgoing edges is equal.
The next three constraints deal with the $y$ variable.
Equation~\eqref{eq:y.1} ensures that each layer has only one phase assigned.
Equations~\eqref{eq:y.2} and \eqref{eq:y.3} impose the monotonicity constraint for all the phases, ensuring that the phases are assigned to layers in ascending order.
Equation~\eqref{eq:z.1} ensures that each operation is assigned precisely one phase.
The three binary variables, $x,y,z$, are linked with Eq.~\eqref{eq:xyz.1}.
The final constraint (Eq.~\eqref{eq:alpha.1}) ensures that the $\alpha$ continuous variable calculates the maximum load of a phase.

The optimal sequence and operations per phase are taken from the $x$ and $z$ binary variables.
From the solution of the \ac{mip} problem, many valuable results can be extracted.
First, the joint sequence and the attributes of the joint and parts for each operation.
This could be used to check the attribute changes (technology) or development (handling, tolerance).
Additionally, the phase/station times are the values we aim to minimize for the \ac{plp} problem.

We utilized the \texttt{PySCIPOpt} library \citep{MaherMiltenbergerPedrosoRehfeldtSchwarzSerrano2016} to solve the problem computationally.

\section{Results}\label{sec:results}
After implementing the computational framework, we evaluated its correctness and performance.
The computational experiments were conducted on two industrial assemblies.

\subsection{Experimental setup}
The first industrial assembly (Assembly 1) consists of 14 parts and 13 joints (Figs.~\ref{fig:ass_1_edges}, \ref{fig:ass_1_nodes}).
This assembly serves as the baseline for verifying the correctness of the framework.
We used Assembly 1 to analyze the sensitivity of the objective function to the user-defined engineering weights ($\mu$) and the time-balancing factor ($\lambda$).

The second assembly (Assembly 2) consists of 15 parts and 17 joints (see \ref{app:ass_2}).
According to Eq.~\eqref{eq:max-edges}, the upper bound for the directed graph edges is 1,114,112 ($\approx1.1\times10^6$), compared to Assembly 1, 53,248 ($\approx 5.3\times10^4$), increasing the complexity of the problem by nearly two orders of magnitude.
Therefore, Assembly 2 provides a suitable test case to evaluate the effectiveness of the geometrical constraints, i.e., \ac{dof} matrices, and the path-guided edge reduction method in maintaining optimal solutions for high-complexity scenarios.

The experiments were conducted on local workstations equipped with an \texttt{Intel Core i5-14500} CPU (2.60 GHz) and 32 GB of RAM.

\subsection{Verification for optimization objectives}

\begin{figure}[h]
    \centering
    \includegraphics[width=0.48\textwidth]{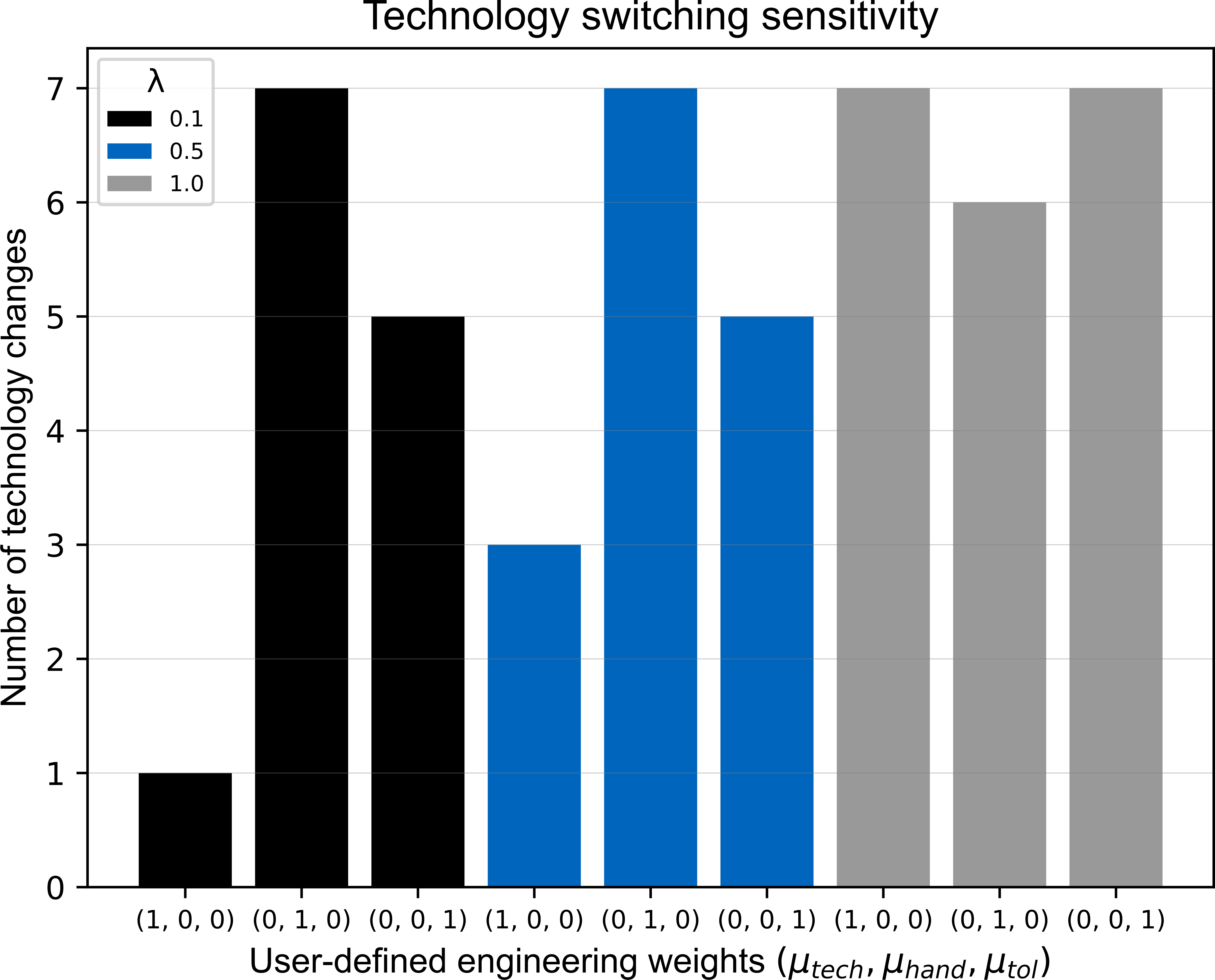}
    \caption{Impact of engineering weights ($\mu$) on technology changes. When \ac{asp} and $\mu_{tech}$ are prioritized, the solver identifies the theoretical minimum of 1 change. On other occasions, there are up to 8 random changes.}
    \label{fig:attr-ch-tech}
\end{figure}

\begin{figure}[h]
    \centering
    \includegraphics[width=0.48\textwidth]{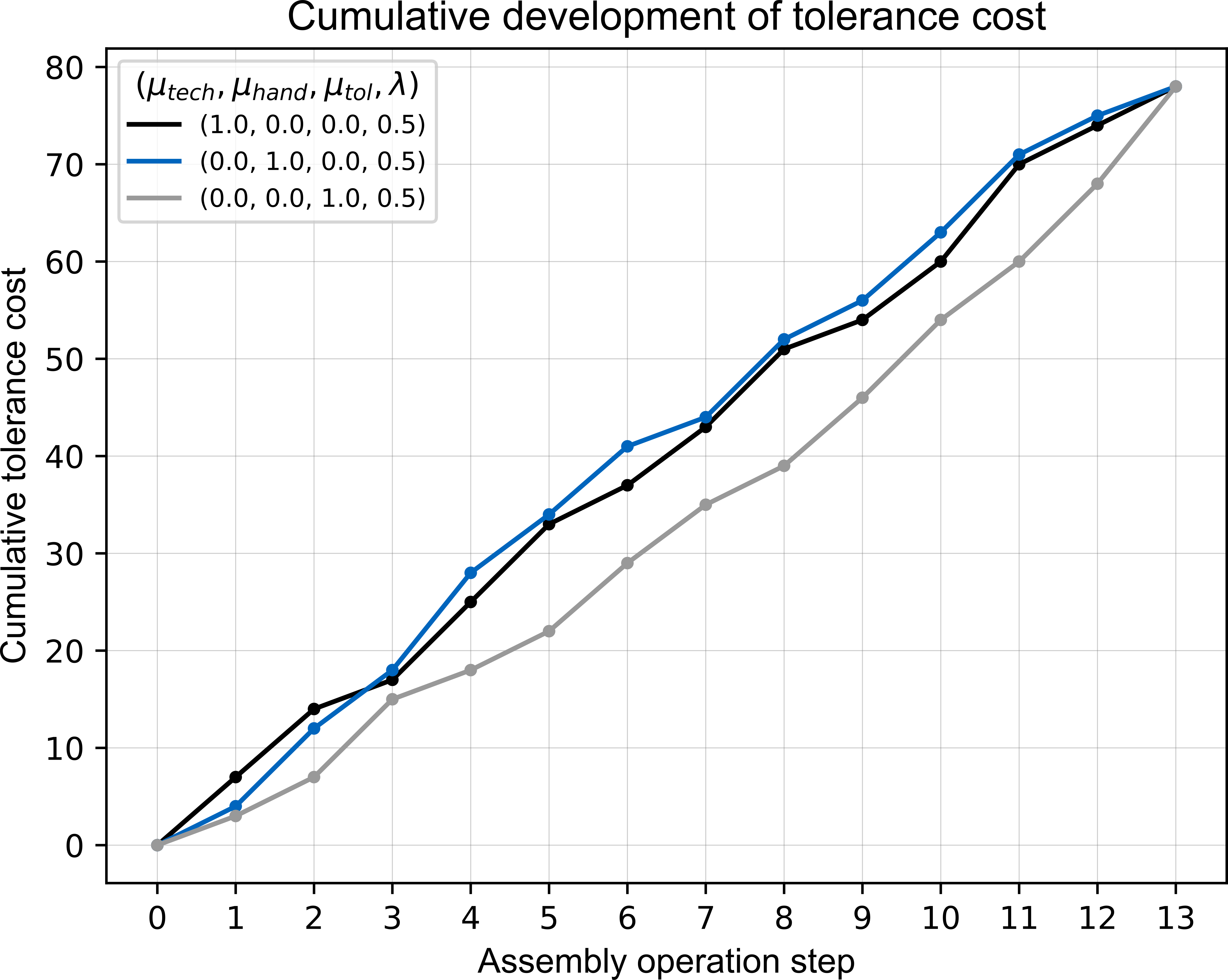}
    \caption{Cumulative development of absolute tolerance costs over the assembly sequence. Prioritizing tolerance ($\mu_{tol}=1.0$, gray line) results in a shallower accumulation curve compared to non-prioritized runs. Note that tolerance is modeled as a normalized cost, where higher values indicate stricter tolerances.}
    \label{fig:attr-dv-tol}
\end{figure}

\begin{figure}[h]
    \centering
    \includegraphics[width=0.48\textwidth]{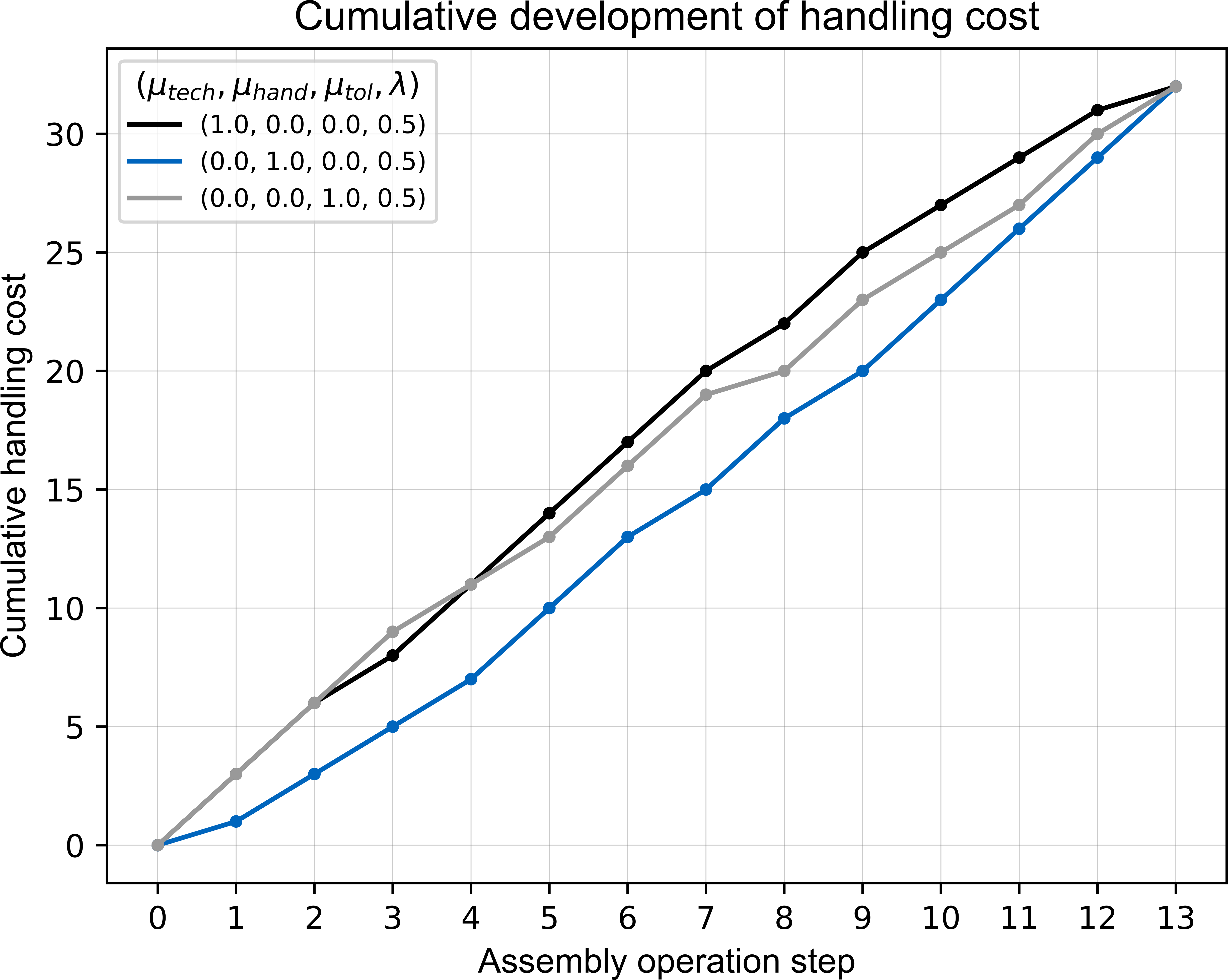}
    \caption{Cumulative development of handling costs. Prioritizing the handling attribute ($\mu_{hand}=1.0$, blue line) minimizes the total handling complexity of the sequence. Note that handling is modeled as a normalized cost, with higher values indicating greater fragility.}
    \label{fig:attr-dv-hand}
\end{figure}

\begin{figure}[h]
    \centering
    \includegraphics[width=0.48\textwidth]{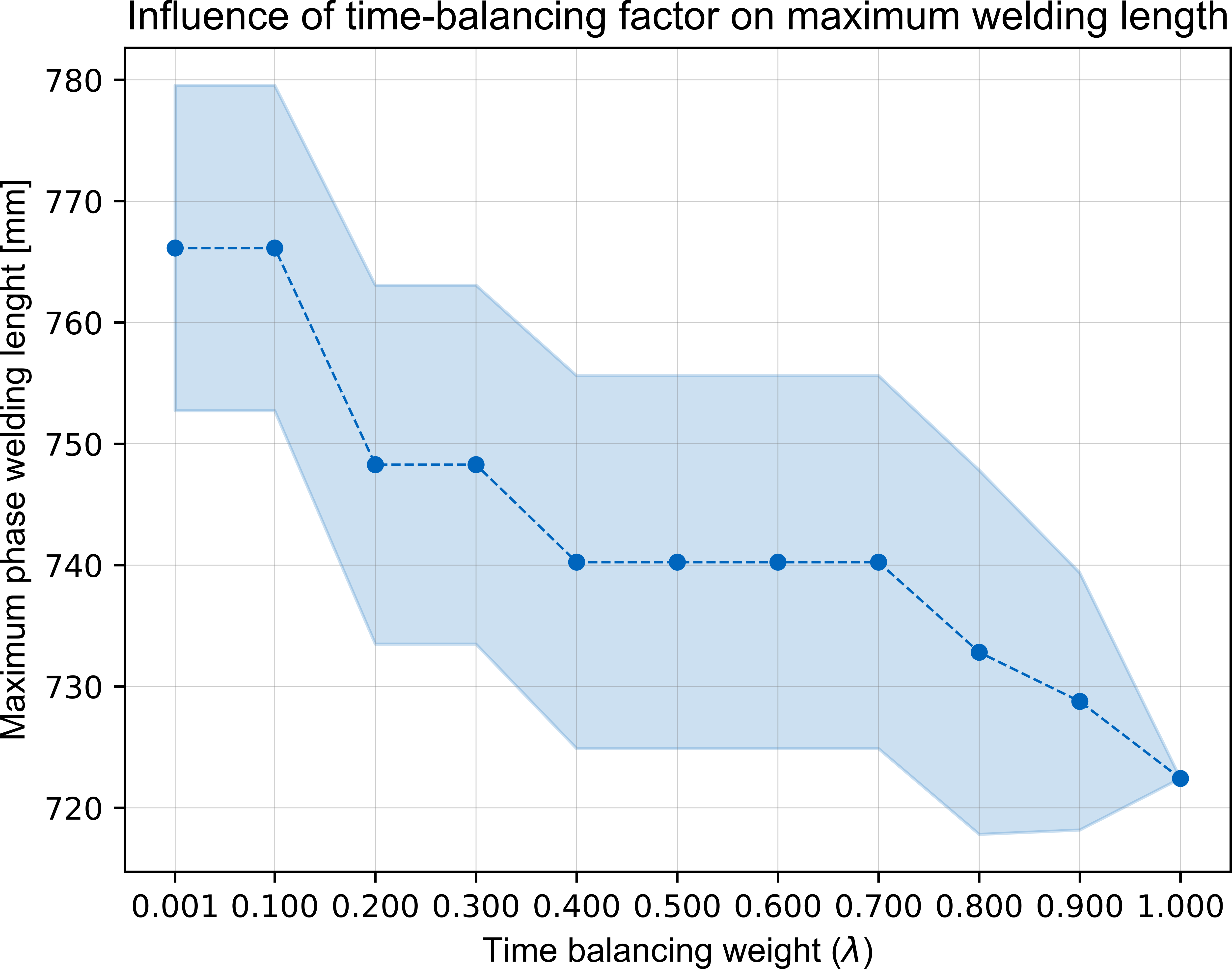}
    \caption{Influence of the time-balancing factor $\lambda$ on the maximum phase welding length, indirectly on the phase time $\alpha$. The shaded region represents $\pm 1$ standard deviation over the $\mu$ configurations. As $\lambda \to 1$, the system effectively prioritizes minimizing cycle time.}
    \label{fig:time-bal-phase}
\end{figure}

This section utilizes Assembly 1 to verify that the proposed objective function $\mathcal{J}$ correctly solves the \ac{asp} and \ac{plp} problems according to user-defined priorities. 
We assess the solver's responsiveness to specific manufacturing constraints by varying the engineering weights ($\mu$) and the time-balancing factor ($\lambda$).
For all validation tests, the number of phases was fixed at $P=3$.

\subsubsection{Sensitivity to engineering weights ($\mu$)}

We analyzed the impact of technology, handling, and tolerance weights on the optimal assembly structure.
In each scenario, one engineering constant was prioritized (set to 1.0), while the other two were ignored (set to 0.0).

Assembly 1 consists of two different technologies ("MAG" and "MAG2"), as shown in Figure \ref{fig:ass_1_edges}.
In that case, the minimum possible number of technology changes is one, i.e., switching from one manufacturing technology to the other.
In addition, three time balancing weight values were used  ($\lambda={0.1, 0.5, 1.0}$).
Figure \ref{fig:attr-ch-tech} shows the number of technology changes presented in the optimal assembly sequence.
The results show that when the technology weight is prioritized ($\mu_{tech}=1.0$), and the time-balancing influence is low to moderate ($\lambda<0.5$), the solver attempts to minimize the number of technology changes.
In the case of $\lambda=0.1$, the minimum possible technology change of one is achieved. 
In contrast, the other scenarios with no technology prioritization ($\mu_{tech}=0.0$) and/or a higher time-balancing factor ($\lambda=1.0$) resulted in up to eight technology changes.
Therefore, the solver correctly penalizes technology switching when $\mu_{tech}$ is prioritized.

In the next step, we evaluated the handling and tolerance constraints.
The optimization objective is to minimize the cumulative value of these attributes by prioritizing sequences with lower part fragility and relaxed tolerance requirements.
We fixed the time-balancing factor at $\lambda=0.5$ to ensure an equal trade-off between the soft engineering constraints and time balancing.
Figures \ref{fig:attr-dv-tol} and \ref{fig:attr-dv-hand} show the cumulative development of the attribute costs over the assembly sequence steps.
In both cases, when the attribute in focus was prioritized ($\mu_{hand}=1$ or $\mu_{tol}=1$), the accumulation curve is notably shallower compared to the other runs.
The smaller area under these curves confirms that the objective function effectively minimizes the specific cumulative cost of the chosen attribute. 

\subsubsection{Sensitivity to time-balancing factor ($\lambda$)}
Finally, we evaluated the influence of the time-balancing factor ($\lambda$) on production line efficiency.
By definition, increasing $\lambda$ prioritizes the time-balancing problem over the \ac{asp} problem, aiming to minimize the maximum phase time ($\alpha$).
We use the assembly welding length as an indirect representation of time.

Figure \ref{fig:time-bal-phase} illustrates the maximum phase welding length across varying $\lambda$ values.
The data shows a downward trend in phase welding length as $\lambda$ approaches 1.0.
Thus, the parameter $\lambda$ serves as a mechanism for regulating the trade-off between phase time and sequence optimality.

\subsubsection{Multi-objective trade-off analysis}

\begin{figure*}[t]
    \centering
    \includegraphics[width=1\textwidth]{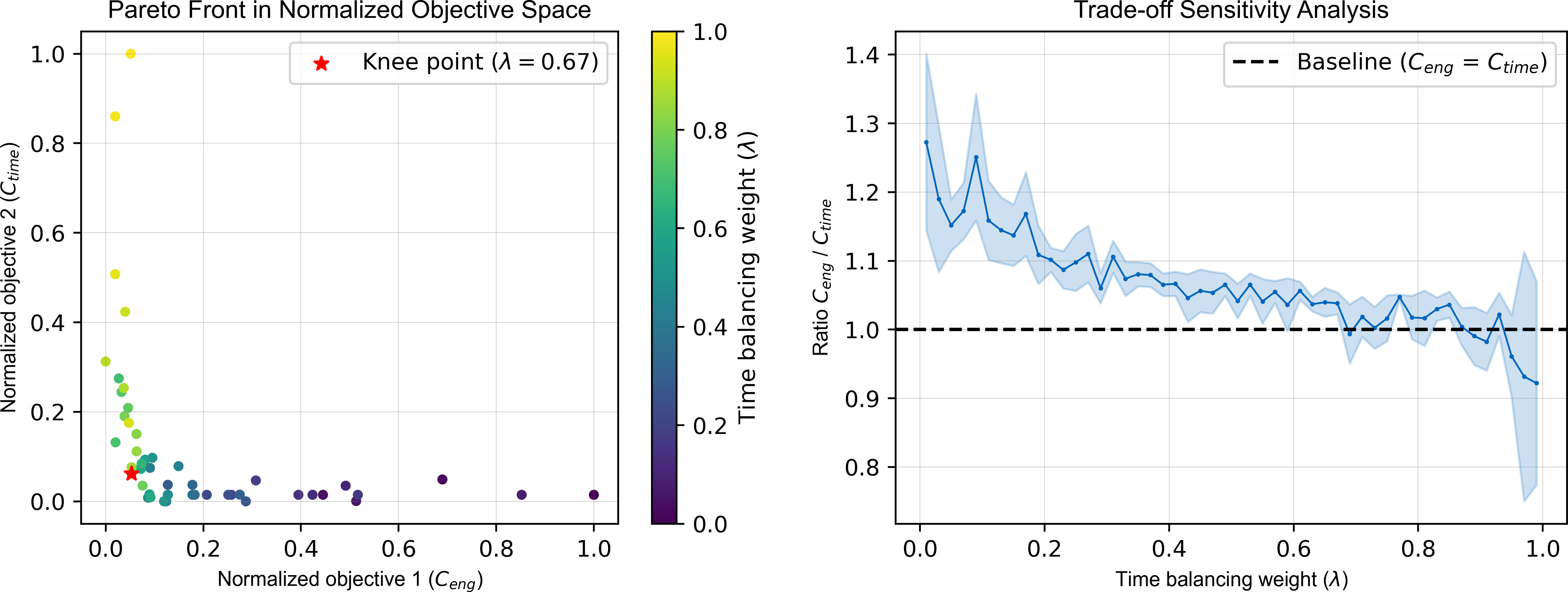}
    \caption{Objectives trade-off evaluation: The left plot shows the Pareto front in normalized objective space, identifying a knee point at $\lambda = 0.67$ where engineering and production line costs reach an optimal equilibrium. 
    The right plot presents a sensitivity analysis of the cost ratio ($C_{eng}/C_{time}$) across varying time-balancing weights $\lambda$, using four representative attribute configurations. 
    The baseline at 1.0 indicates the transition from \ac{asp}-focused to \ac{plp}-focused solutions.}
    \label{fig:pareto_plots}
\end{figure*}

In the final verification step, the trade-off between the two primary objectives was evaluated.
Specifically, the engineering cost, $\mathcal{C}_{eng}=\sum_{e \in E}(w_e x_e)$, and the time-balancing cost, $\mathcal{C}_{time}=c \alpha$, (see equations \ref{eq:obj_func} and \ref{eq:obj}), were calculated for $\lambda$ values in $(0,1)$.
The resulting Pareto Front (Figure~\ref{fig:pareto_plots}, left) illustrates the distribution of non-dominated solutions in the normalized objective space, with a distinct knee point identified at $\lambda = 0.67$.
Both objectives are min--max normalized to $[0,1]$ over the computed non-dominated set, so that the origin of Figure~\ref{fig:pareto_plots} (left) corresponds to the \emph{utopia point}: the generally unattainable solution that simultaneously minimizes $\mathcal{C}_{eng}$ and $\mathcal{C}_{time}$.
The knee point is identified as the non-dominated solution with the smallest Euclidean distance to this utopia point,
\begin{equation}
    \lambda^{*} = \arg\min_{\lambda}
    \sqrt{\widehat{\mathcal{C}}_{eng}(\lambda)^{2} + \widehat{\mathcal{C}}_{time}(\lambda)^{2}},
    \label{eq:knee}
\end{equation}
where $\widehat{\mathcal{C}}$ denotes the normalized cost.
This criterion weights both objectives equally and depends on the chosen normalization; we use min--max scaling over the computed front.
This point represents the optimal equilibrium in which both objectives are minimized without one being disproportionately penalized.

To assess the robustness of these solutions, a trade-off sensitivity analysis was conducted (Figure~\ref{fig:pareto_plots}, right), incorporating four representative engineering weight ($\mu$) configurations: equal weights, and technology-centric, handling-centric, and tolerance-centric.
The analysis demonstrates that across all $\lambda$ ranges, the cost ratio remains within an average $30\%$ deviation from the baseline.
While the ratio naturally increases when optimizing for a single objective (near $\lambda=0 $ or $1$), the overall convergence verifies that the equal contribution factor $c$ effectively balances both domains.

The sensitivity plot reveals three distinct optimization regions relative to the baseline ($y = 1.0$):
\begin{enumerate}
    \item \textbf{\ac{asp}-dominant area ($\lambda < 0.4$):} The $\mathcal{C}_{eng}$/$\mathcal{C}_{time}$ ratio remains consistently above the baseline.
    Despite stochastic fluctuations (due to $\mu$ configurations), a clear downward trend is visible as $\lambda$ increases.
    This range is recommended for users prioritizing \ac{asp}-specific constraints, as these attributes exert a $>10\%$ greater influence than time-balancing costs.
    
    \item \textbf{Balanced equilibrium ($0.4 \leq \lambda \leq 0.9$):} The mean ratio converges toward the baseline, with fluctuations narrowing to within a $<10\%$ margin.
    This stability reflects the framework's ability to harmonize competing objectives.
    Notably, the trend crosses the threshold near the knee point ($\lambda = 0.67$), and remains in close proximity to the baseline until $\lambda = 0.9$.
    
    \item \textbf{\ac{plp}-dominant area ($\lambda > 0.9$):} As $\lambda$ approaches $1.0$, the ratio dips below the baseline, indicating that production line efficiency becomes the primary driver.
    The convergence in this area demonstrates the framework's ability to consistently minimize manufacturing station time.
    In combination with the results in Figure~\ref{fig:time-bal-phase}, for \ac{plp}-critical problems, users should use a value of $\lambda > 0.9$.
\end{enumerate}

The alignment between the Pareto knee point and the baseline crossing verifies that the framework provides three distinct regions that govern competing industrial priorities and adapt to a wide spectrum of manufacturing scenarios.

\subsection{Computational efficiency and scalability}
\label{subsec:comp-eff}

This section evaluates the framework's performance on both industrial assemblies, with an emphasis on the complex Assembly 2, focusing on the impact of physical constraints (i.e., collision checks) and deterministic edge-reduction methods on computational time and solution quality.

\subsubsection{Geometrical constraints effect}

For this assembly, the solution space is significantly larger, with a theoretical maximum of approximately $1.1\times 10^6$ edges (Eq.~\eqref{eq:max-edges}).
Enforcing Single-Piece Flow alone reduces this to $244{,}029$ edges.
Enabling the collision check using \ac{dof} matrices during directed graph creation reduced the number of edges further by $45\%$, leaving the $134{,}216$-edge graph used throughout this section.
This \ac{dof} reduction step significantly reduces the graph size before the \ac{plp} phase begins and is essential for both handling complex assemblies and geometrical feasibility.

\subsubsection{Path-guided edge reduction}
\label{subsec:scalability}

We evaluate the deterministic reduction along three axes: which subgraph generation strategy yields the most efficient subgraph, how quickly the reduced \ac{mip} converges to the full optimum, and how its solving time compares to that of the full-graph \ac{mip}, when its calculation is possible.
For the $k$-value determination, we used a set of values in the range of $k \in \{1, 2, 3, 4, 6, 8,\dots, 256, 512, 1024, 2048\}$.
Note that $k$ is used as an internal path generation parameter, since the number of individual discovered paths and the subgraph percentage are dependent on the full directed graph.
Therefore, the subgraph percentage stop value is used as an assembly-agnostic criterion.

\begin{figure}[h]
    \centering
    \includegraphics[width=0.48\textwidth]{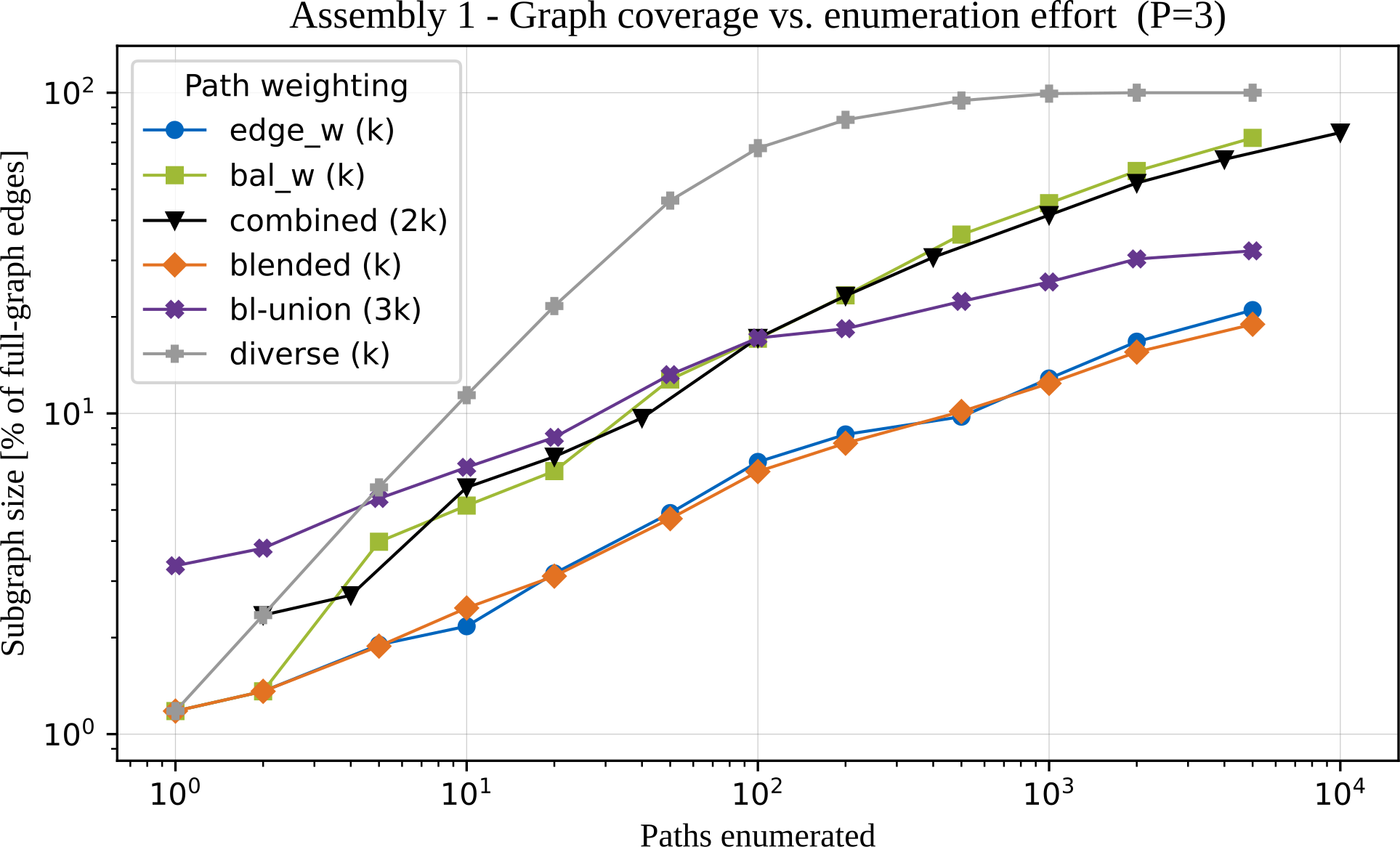}
    \caption{Graph coverage versus enumeration effort for the six path-weighting strategies (Assembly 1, $P=3$). The vertical axis shows the resulting subgraph size as a percentage of the full directed graph's edges. The horizontal axis shows the number of enumerated paths. The \textit{diverse} strategy attains the highest coverage for a given number of paths.}
    \label{fig:k_vs_subass}
\end{figure}

\begin{figure*}[t]
    \centering
    \includegraphics[width=1\textwidth]{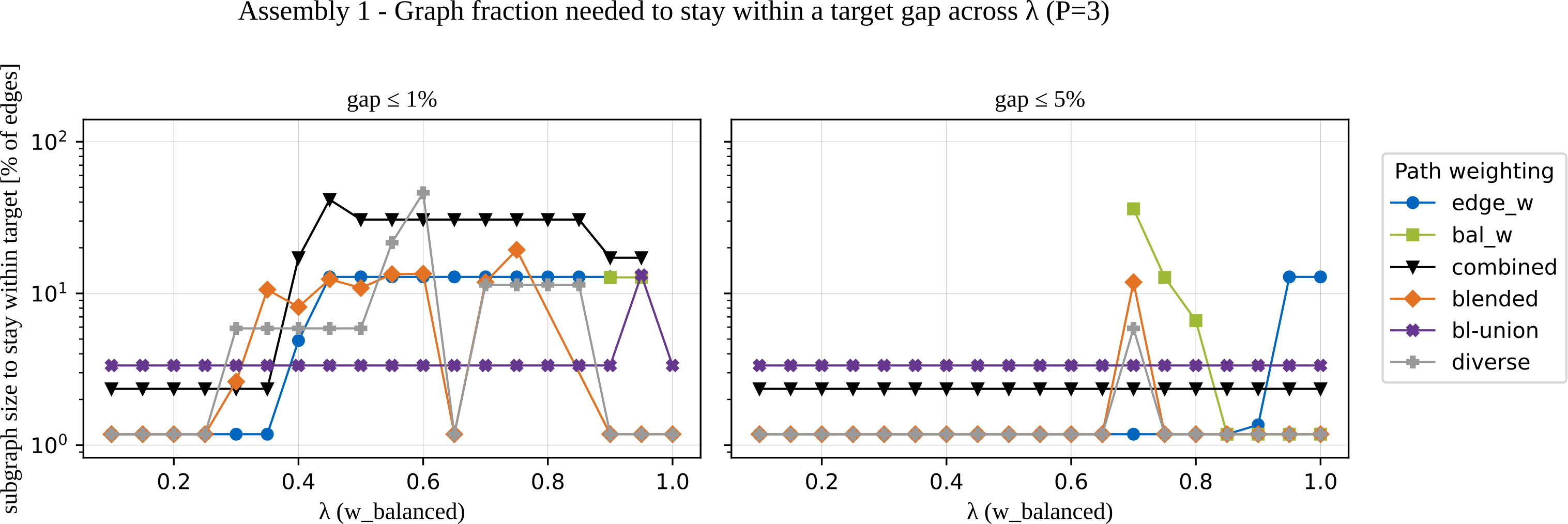}
    \caption{Comparison of the six path-weighting strategies (Assembly 1, $P=3$).
    For each strategy, the plots show the subgraph size required to bring the reduced-\ac{mip} objective within a target gap of the full-graph optimum, as a function of the time-balancing factor $\lambda$. Left: target gap $\leq 1\%$. Right: target gap $\leq 5\%$. Lower is better.}
    \label{fig:min_k_for_subgraph}
\end{figure*}

\textbf{Enumeration strategies:} Figure~\ref{fig:k_vs_subass} shows the graph coverage of the six strategies compared to the number of enumerated paths.
The diverse method coverage is higher for the same number of paths than for all other methods.
Furthermore, figure~\ref{fig:min_k_for_subgraph} compares the strategies by the subgraph fraction each needs to bring the reduced-\ac{mip} objective within a target gap of the full optimum, across $\lambda$.
The engineering-weight enumeration (\texttt{edge\_w}) is the baseline.
\texttt{Bl-union} and \texttt{diverse} reach the target at the smallest fractions across the whole $\lambda$ range.
The same trend was observed on the same experiments with $P=2,4,5$.

The two dominant strategies have weaknesses that are complementary: \texttt{bl-union}'s coverage saturates, i.e., the number of independent paths stops growing as more are enumerated, while \texttt{diverse} grows without bound, but its penalty ordering can skip high-quality paths.
Therefore, we implemented an adaptive strategy that combines them, inheriting the strengths of both.

\begin{figure}[h]
    \centering
    \includegraphics[width=0.48\textwidth]{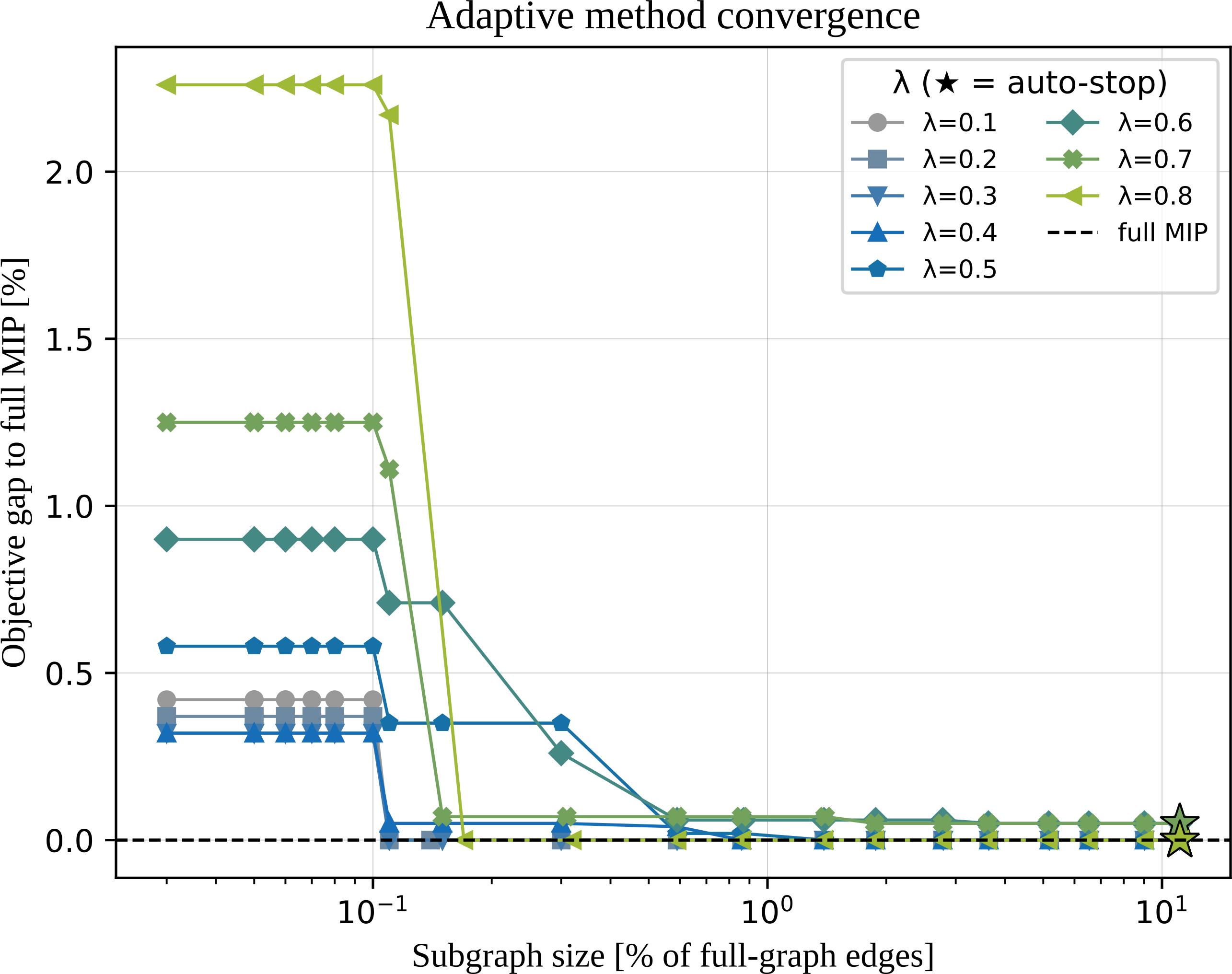}
    \caption{Convergence of the adaptive method for Assembly 2 ($P=3$) across various $\lambda$: objective gap to the full-graph optimum as a function of subgraph size.
    The star marks the auto-stop, i.e.\ the point at which enumeration reaches the $10\%$ ceiling.
    The gap closes well before this ceiling; Table~\ref{tab:speedup} reports the smallest subgraph at which the optimum is attained.
    }
    \label{fig:adapt_conv_ass_2}
\end{figure}

\textbf{Convergence to the full optimum:} Figure~\ref{fig:adapt_conv_ass_2} shows, for each $\lambda$, the objective gap of the reduced \ac{mip} to the full-graph optimum of Assembly~2 as a function of the subgraph fraction, for the adaptive method.
The gap falls sharply and reaches (near) zero within a small fraction of the graph.
The markers denote the enumeration auto-stop.

\textbf{Solving time versus the full \ac{mip}:}
Table~\ref{tab:speedup} compares the overall time of the deterministic adaptive reduction against the full-graph \ac{mip} on Assembly~2 ($P=3$), across the time-balancing factor~$\lambda$.
The full-\ac{mip} time is the solve time on the complete $134{,}216$-edge graph, where the reduced time includes the total time of generating the subgraph with the adaptive method (path enumeration and the \ac{mip} solve) at the smallest subgraph that attains the closest solution to the full-graph optimum.
The reduction reaches the exact optimum at below $2\%$ of the graph for six of the eight $\lambda$ values, with the largest gains where the full \ac{mip} is hardest: at $\lambda=0.8$, a full solve of $\sim\!281$~min is reproduced exactly in $11$~s, a $1514\times$ speedup.
For $\lambda\in\{0.6,0.7\}$, the optimum is approached to within $0.05\%$ but not closed within the tested budget.
Note that Table~\ref{tab:speedup} reports the smallest subgraph attaining the optimum, whereas the auto-stop in Figure~\ref{fig:adapt_conv_ass_2} is a conservative $10\%$ ceiling.
Since the optimum is captured at a much smaller fraction, enumerating up to the ceiling incurs computational overhead without improving solution quality, so the percentage budget should be set conservatively low.

\begin{table}[t]
\centering
\caption{Deterministic adaptive reduction versus the full \ac{mip} on Assembly~2 ($P=3$, $134{,}216$ edges). $t_{\text{full}}$: full-graph \ac{mip} solve time.
Subgraph: the smallest fraction of $E$ attaining the best solution compared to the full-graph optimum.
$t_{\text{red}}$: subgraph generation with adaptive method $+$ solve time at that fraction. 
Gap: objective deviation from the full optimum.}
\label{tab:speedup}
\begin{tabular}{@{}lrrrrr@{}}
\toprule
$\lambda$ & $t_{\text{full}}$ [s] & Subgraph [\%] & $t_{\text{red}}$ [s] & Speedup & Gap [\%] \\
\midrule
0.1 & 276.8    & 0.11 & 17.8 & $16\times$    & 0.00 \\
0.2 & 406.2    & 0.11 & 17.4 & $23\times$    & 0.00 \\
0.3 & 1\,951.3 & 0.11 & 15.5 & $126\times$   & 0.00 \\
0.4 & 9\,087.2 & 0.86 & 22.6 & $402\times$   & 0.00 \\
0.5 & 6\,006.0 & 1.40 & 39.7 & $151\times$   & 0.00 \\
0.6 & 10\,837.4 & 3.63 & 133.1 & $81\times$   & 0.05\\
0.7 & 11\,462.4 & 1.87 & 60.7  & $189\times$  & 0.05\\
0.8 & 16\,861.0 & 0.17 & 11.1 & $1\,514\times$ & 0.00 \\
\bottomrule
\end{tabular}
\end{table}

\begin{figure*}[t]
    \centering
    \includegraphics[width=1.0\textwidth]{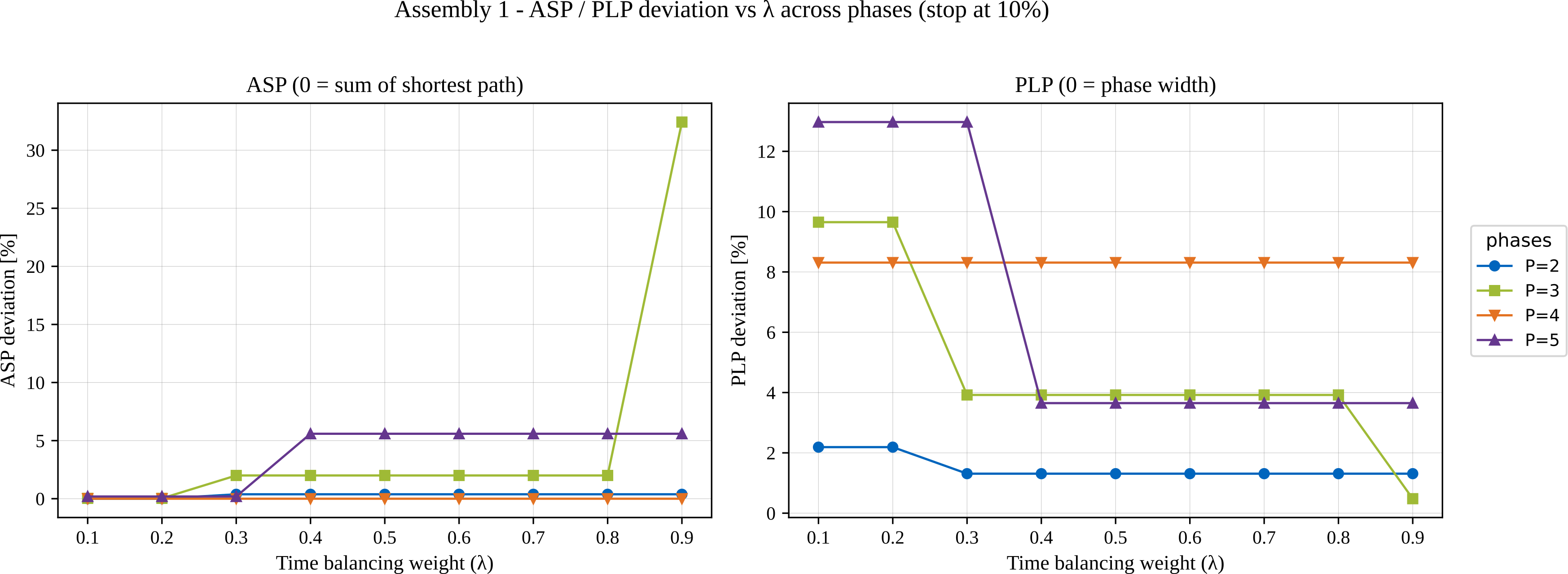}
    \caption{Assembly 1: Deviation from lower bounds for \ac{asp} (sum of the edge weights of the shortest path) and \ac{plp} (phase width).
    Subgraph stop criterion: $10\%$.
    }
    \label{fig:lamb_v_p_ass_1}
\end{figure*}

\begin{figure*}[t]
    \centering
    \includegraphics[width=1.0\textwidth]{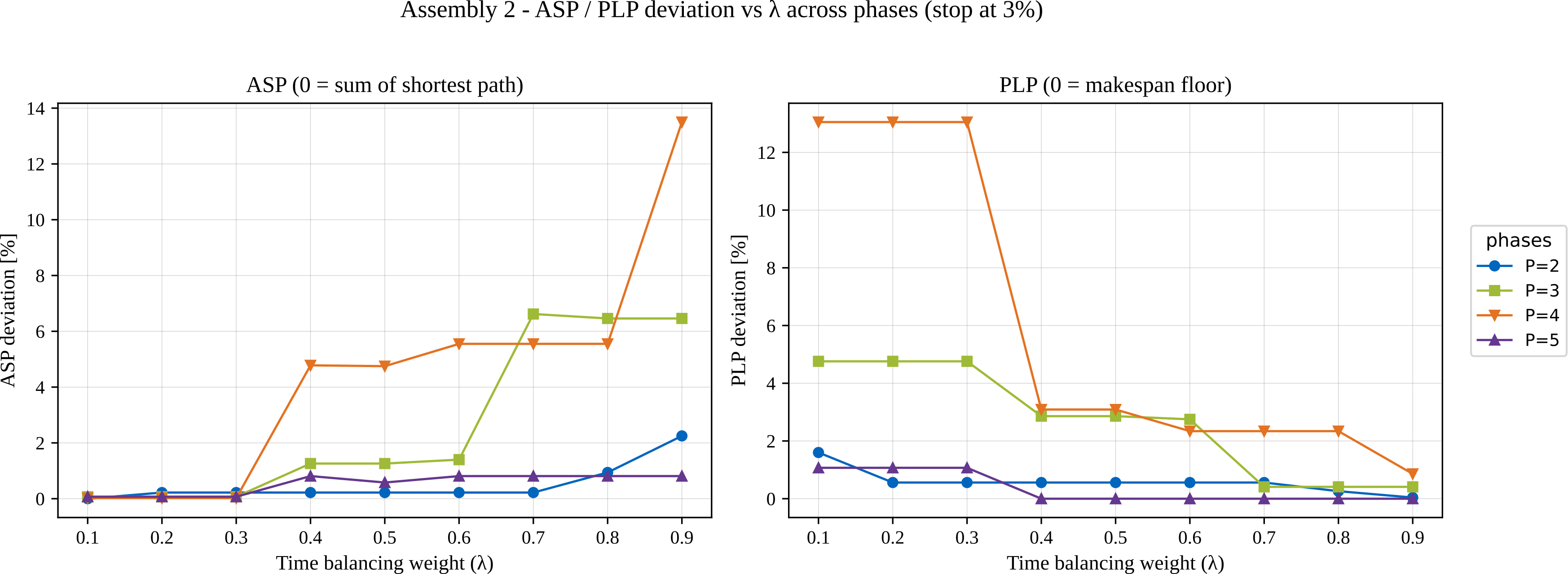}
    \caption{Assembly 2: Deviation from lower bounds for \ac{asp} (sum of the edge weights of the shortest path) and \ac{plp} (phase width).
    Subgraph stop criterion: $3\%$.
    }
    \label{fig:lamb_v_p_ass_2}
\end{figure*}

\textbf{Optimality references and quality across phases:} The previous section compared the reduced \ac{mip}'s solution to the full-\ac{mip}.
The effectiveness of the proposed framework should be tested on cases where the \ac{mip} is infeasible.
In this section, the reduced problem solution is tested against theoretical optimal values covering both \ac{asp} and \ac{plp} problems.
Two lower bounds frame the solution's optimality: (i) an \ac{asp} bound, the sum of the engineering-weight shortest-path edges (Eq.~\eqref{eq:edge_weight}), and (ii) a \ac{plp} bound, the makespan floor $\tfrac{1}{P}\sum_{o\in O} t_o$ (the ideal phase width). 
Figures~\ref{fig:lamb_v_p_ass_1} and~\ref{fig:lamb_v_p_ass_2} report the \ac{asp} and \ac{plp} deviations from these bounds across $\lambda$ and the number of phases $P$, both of which remain small over the tested range, with the residual \ac{plp} deviation reflecting that the two bounds are decoupled and generally not simultaneously attainable.
An engineer can run multiple low-cost problems and compare the tradeoff across the $\lambda$ and $\mu$ values.

\textbf{Industrial validation:} For a final validation test, we compared the best time balancing solution for Assembly~2 for initial factory line planning design, against an industrial baseline.
The industrial baseline was designed using the welding length as a proxy for time-balancing to ensure a fair initial production-planning comparison.
The framework's optimized maximum phase length for three stations is $900$~mm at $\lambda=0.8$.
This represents a $19.6\%$ reduction compared to the experience-based industrial baseline designed for five stations, which has a maximum phase length of $1120$~mm (joints $ j_4$ and $ j_5$; see Appendix~C).
Because the deterministic reduction reproduces the exact full-\ac{mip} optimum (Table~\ref{tab:speedup}), this $19.6\%$ improvement is preserved under the reduction rather than degrading with the graph size.

\section{Discussion}

The results demonstrate that the PyCAALP framework successfully harmonizes the competing priorities of \ac{asp} and \ac{plp}. 
A critical factor in this success is the enforcement of geometric constraints.
By utilizing \ac{dof} matrix-based collision detection, the framework proactively prunes approximately $45\%$ of infeasible operations before the optimization step.
This bridges the gap between purely theoretical sequence generation and physically viable manufacturing.

Furthermore, scaling such unified problems for complex assemblies often leads to a combinatorial explosion. 
The proposed path-guided edge reduction addresses this by solving the \ac{mip} on a deterministic subgraph assembled from complete, high-quality paths.
Because every retained path is complete, the reduced \ac{mip} is always feasible and its optimum bounds the full-graph optimum.
The reduction, therefore, carries a feasibility-and-bound guarantee rather than relying on a stochastic safeguard.
On Assembly~2, the adaptive strategy reproduces the exact full-\ac{mip} optimum at below $2\%$ of the directed graph for six of the eight tested $\lambda$ values, with the largest gains arising where the full \ac{mip} is hardest and the optimum is captured early, reaching $1514\times$ (from $\sim\!281$~min to $11$~s) at $\lambda=0.8$ (Table~\ref{tab:speedup}).
The \ac{plp} solution was validated against an experience-based industrial baseline: the three-station solution achieved a $19.6\%$ reduction in maximum phase length ($900$~mm) versus the manual five-station design ($1120$~mm).

Finally, the sensitivity analysis confirms that the unified objective function enables users to precisely control the manufacturing setup. 
By adjusting the time-balancing factor ($\lambda$) and engineering weights ($\mu$), engineers can actively prioritize specific assembly risks (i.e, technology switching, fragility, tolerance accumulation) against production line efficiency.
However, it should be noted that the current \ac{mip} formulation optimizes for a pre-defined, fixed number of workstations.
With the speedup option, however, it is straightforward to compute solutions for different numbers of workstations, which also allows for estimating dynamic line-scaling options.
The framework further works under the constraint of a strictly sequential assembly process governed by the single-piece flow heuristic.

By operating on only one active subassembly at a time (single-piece-flow heuristic), the framework inherently restricts parallel execution.
Although we used a fixed problem setup for our work (Single-Piece Flow), the framework can accommodate parallel assemblies (see Appendix B)
While extending the model to support multiple simultaneous subassemblies is theoretically feasible, it would substantially increase computational complexity. 
As a practical workaround for parallel stations with a priori known subassemblies, the optimization could be solved individually for each subassembly and balanced post-process. 
Nevertheless, the current implementation cannot handle parallel tasks or parallel workstations, which remains a limitation when modeling the dynamics of high-volume, flexible manufacturing environments. 
However, the proposed framework and methods are not limited to the single-piece-flow heuristic and provide a generic basis for extension.

Using the welding length as the operating time is a modeling choice that aligns with an industrial approach for initial factory planning and design.
The formulation itself is agnostic: $t_o$ in Eq.~\eqref{eq:alpha.1} is an arbitrary per-operation cost.
Incorporating sequence-dependent terms (tool approach, repositioning) could be handled by a directed graph time-weight calculation, where $t_e$ will substitute $t_o$.
These modifications will not change the \ac{mip} structure.
Furthermore, the inter-station transfer time is not a structural limitation: a per-edge transfer cost can be added to the $\alpha$ constraint whenever an edge crosses a station boundary, attributing it to that station's load. 
Because the industrial baseline was itself balanced on welding length, the $1120$\,mm versus $900$\,mm comparison is like-for-like and isolates the benefit of the optimization under a common metric. 
It does not, and is not intended to, validate welding length as a proxy for measured cycle time. 
That validation remains an open direction.

The dominant computational cost of the framework is the \ac{mip} solving step, not the weighted directed graph generation. 
It is therefore the \ac{mip} that bounds the practical assembly size. 
Once the reduction is applied, however, this balance inverts: at the small subgraph fractions of Table~\ref{tab:speedup}, the path enumeration, not the \ac{mip} solve, accounts for most of the reported time, so further acceleration should target the enumeration step.
The number of directed graph edges shows a worst-case bound (Eq.~\ref{eq:max-edges}), $5.3\times10^{4}$ at $J{=}13$, $1.1\times10^{6}$ at $J{=}17$, and $1.0\times10^{7}$ at $J{=}20$, but this describes only the worst case, not the difficulty of the resulting \ac{mip}.
The effective problem size is assembly-dependent: single-piece-flow eliminates disconnected cutsets and the \ac{dof} collision check prunes, an additional $\sim$$45\%$ of operations (Assembly 2 case), while the solving time of the remaining \ac{mip} is governed by the specific graph structure rather than by the joint count alone.
For this reason, a general theoretical maximum tractable joint count cannot be stated, but rather it has to be computed per assembly from its actual combinations.
In practice, small assemblies (Assembly~1) are solved directly, larger ones (Assembly~2) require the deterministic path-guided edge reduction to reach minute-scale solving times, and assemblies beyond this range can be decomposed into subassembly modules to be solved independently. 
Finally, the framework operates on an arbitrary connected part graph $G_{part}$ and imposes no restriction on its topology: the two test assemblies differ not only in size (13 vs.\ 17 joints) but also in structure, with Assembly~1 a near-linear weldment and Assembly~2 a branched, non-linear structure (Fig.~C3), so the approach is not limited to linear topologies, and can be used for different types of manufacturing parts.

Beyond facility layout constraints, the framework's algorithmic design presents two additional characteristics that warrant discussion: one limitation and one property of the reduction. 
First, the sequence generation process is based on a disassembly approach, which may struggle to capture nuanced forward-assembly operations, such as installing temporary fixtures, performing intermediate calibration steps, or applying consumables like lubricants and adhesives. However, assuming these specific tasks can be predefined, those additional operations can be modeled as additional nodes and joints in the framework. An inclusion regarding collision control is straightforward in this regard as well.
Second, the cutset-complexity reduction is deterministic: the reduced subgraph is a fixed function of the enumeration weights and the size budget, and it preserves the feasibility-and-bound guarantee for every instance.



\section{Conclusions}
\label{sec:conclusions}
This work presents PyCAALP, a unified, open-source computational framework for automated Assembly Sequence Planning (ASP) and Production Line Planning (PLP).
By integrating graph-based modeling with Mixed-Integer Programming, the framework successfully bridges the gap between product design parameters and production line efficiency.
Verification on two industrial assemblies with differing sizes and topologies confirmed that it handles high combinatorial complexity and produces near-optimal solutions in minutes of computation for the tested cases.
Three properties underpin this result: \ac{dof}-based collision detection that removes geometrically infeasible operations before optimization, a deterministic path-guided edge reduction that reproduces the exact or near-exact \ac{mip} optimum at a small fraction of the directed graph with a feasibility-and-bound guarantee, and a unified objective that lets engineers weight assembly-quality (\ac{asp}) against line-efficiency (\ac{plp}) priorities through user-defined parameters.
Because the formulation operates on an arbitrary connected part graph and treats the per-operation cost as a generic input, the approach is not specific to the arc-welded assemblies studied here and, in principle, extends to other joining processes and assembly topologies.

Several directions remain open. 
The current formulation assumes a fixed number of sequential, single-piece-flow stations. 
Supporting parallel stations and variable station counts would broaden its applicability to high-volume, flexible production lines. 
The operation-time model can be enriched with sequence-dependent and inter-station transfer terms, and validated against measured cycle times rather than a geometric proxy. 
Finally, automatic data extraction directly from raw CAD models will further reduce manual input requirements.
The open-source release of this framework aims to promote further collaboration and adoption across both industrial and academic manufacturing domains.

\section*{Acknowledgements}\label{sec:outlook}
The authors would like to thank the Bavarian State Ministry of Economic Affairs, Regional Development and Energy (StMWi) for funding this project (number DIK0280/04).

\bibliography{References.bib}

@article{laperrire_al_1996_GAPP_a_generative_assembly_process_planner,
  title={GAPP: A generative assembly process planner},
  author={Luc Laperri{\`e}re and Hoda Elmaraghy},
  journal={Journal of Manufacturing Systems},
  year={1996},
  volume={15},
  pages={282-293},
  url={https://api.semanticscholar.org/CorpusID:109708282}
}

@article{Zhu_Xu_Wang_Yang_Fan_2023_Graph_based_assembly_sequence_planning_algorithm_with_feedback_weights, 
  title={Graph-based assembly sequence planning algorithm with feedback weights}, 
  volume={125}, 
  DOI={10.1007/s00170-022-10639-9}, 
  number={7–8}, 
  journal={The International Journal of Advanced Manufacturing Technology}, 
  author={Zhu, Xiaojun and Xu, Zhigang and Wang, Junyi and Yang, Xiao and Fan, Linlin}, 
  year={2023}, 
  month={Feb}, 
  pages={3607–3617},
  url={https://link.springer.com/article/10.1007/s00170-022-10639-9}
}

@article{Shi_Tian_Ma_Wu_Gu_2024_A_knowledge_graph_based_structured_representation_of_Assembly_process_planning_combined_with_Deep_Learning, 
 title={A knowledge graph–based structured representation of Assembly process planning combined with Deep Learning},
 volume={133}, 
 DOI={10.1007/s00170-024-13785-4}, 
 number={3–4}, 
 journal={The International Journal of Advanced Manufacturing Technology}, 
 author={Shi, Xiaolin and Tian, Xitian and Ma, Liping and Wu, Xv and Gu, Jianguo}, 
 year={2024}, 
 month={Jun}, 
 pages={1807–1821},
 url={https://link.springer.com/article/10.1007/s00170-024-13785-4}
}

@article{Shi_Xiaolin_Tian_al_2022_Knowledge_graph_based_assembly_knowledge_towards_complex_product_assembly_process,
  AUTHOR = {Shi, Xiaolin and Tian, Xitian and Gu, Jianguo and Yang, Fan and Ma, Liping and Chen, Yun and Su, Tianyi},
  TITLE = {Knowledge Graph-Based Assembly Resource Knowledge Reuse towards Complex Product Assembly Process},
  JOURNAL = {Sustainability},
  VOLUME = {14},
  YEAR = {2022},
  NUMBER = {23},
  ARTICLE-NUMBER = {15541},
  URL = {https://www.mdpi.com/2071-1050/14/23/15541},
  ISSN = {2071-1050},
  DOI = {10.3390/su142315541}
}

@article{xia_lu_lu_al_2024_Semantic_knowledge_driven_A_GASeq_A_dynamic_graph_learning_approach_for_assembly_sequence_optimization,
title = {Semantic knowledge-driven A-GASeq: A dynamic graph learning approach for assembly sequence optimization},
journal = {Computers in Industry},
volume = {154},
pages = {104040},
year = {2024},
issn = {0166-3615},
doi = {https://doi.org/10.1016/j.compind.2023.104040},
url = {https://www.sciencedirect.com/science/article/pii/S0166361523001902},
author = {Luyao Xia and Jianfeng Lu and Yuqian Lu and Wentao Gao and Yuhang Fan and Yuhao Xu and Hao Zhang},
}

@article{jing_zhou_zhang_al_2024_XMKR_Explainable_manufacturing_knowledge_recommendation_for_collaborative_design_with_graph_embedding_learning,
title = {XMKR: Explainable manufacturing knowledge recommendation for collaborative design with graph embedding learning},
journal = {Advanced Engineering Informatics},
volume = {59},
pages = {102339},
year = {2024},
issn = {1474-0346},
doi = {https://doi.org/10.1016/j.aei.2023.102339},
url = {https://www.sciencedirect.com/science/article/pii/S1474034623004676},
author = {Yanzhen Jing and Guanghui Zhou and Chao Zhang and Fengtian Chang and Hairui Yan and Zhongdong Xiao},
}

@article{Laperrière_ElMaraghy_1994_GAPP2_Assembly_sequences_planning_for_Simultaneous_Engineering_Applications,
 title={Assembly sequences planning for Simultaneous Engineering Applications},
 volume={9}, DOI={10.1007/bf01751121}, 
 number={4}, 
 journal={The International Journal of Advanced Manufacturing Technology}, 
 author={Laperrière, Luc and ElMaraghy, Hoda A.}, 
 year={1994}, 
 month={Jul}, 
 pages={231–244},
 url={https://link.springer.com/article/10.1007/BF01751121}
}

@article{line-bal-review,
title = {Assembly line balancing: What happened in the last fifteen years?},
journal = {European Journal of Operational Research},
volume = {301},
number = {3},
pages = {797-814},
year = {2022},
issn = {0377-2217},
doi = {https://doi.org/10.1016/j.ejor.2021.11.043},
url = {https://www.sciencedirect.com/science/article/pii/S0377221721009942},
author = {Nils Boysen and Philipp Schulze and Armin Scholl},
}

@article{li.zhi.2024,
  title = {Chance-constrained stochastic assembly line balancing with branch, bound and remember algorithm},
  author = {Li, Zhi and Sikora, Christian G. S. and Kucukkoc, Ibrahim},
  journal = {Annals of Operations Research},
  volume = {335},
  pages = {491--516},
  year = {2024},
  publisher = {Springer},
  doi = {10.1007/s10479-023-05809-1},
  url = {https://doi.org/10.1007/s10479-023-05809-1},
}

@article{Ahmad.2024,
title={A Comprehensive Review: Analysing the Pros and Cons of Assembly Line Balancing Methods},
volume={44}, 
DOI={10.37934/araset.44.2.7288},
number={2},
journal={Journal of Advanced Research in Applied Sciences and Engineering Technology},
author={Aida Husna Ahmad and Osman Shahrul Azmir and Saliza Azlina Osman and Mohd Faris Afiq Mohd Azhar and Mohd Hazrein Jamaludin and Haziq Asyraaf Abu Bakar and Muhammad Izzuan Abd Rahman and Tay Sin Kiat},
year={2024},
month={Apr.},
pages={72–88}
}

@incollection{MaherMiltenbergerPedrosoRehfeldtSchwarzSerrano2016,
  author = {Stephen Maher and Matthias Miltenberger and Jo{\~{a}}o Pedro Pedroso and Daniel Rehfeldt and Robert Schwarz and Felipe Serrano},
  title = {{PySCIPOpt}: Mathematical Programming in Python with the {SCIP} Optimization Suite},
  booktitle = {Mathematical Software {\textendash} {ICMS} 2016},
  publisher = {Springer International Publishing},
  pages = {301--307},
  year = {2016},
  note = {Version 5.0.1},
  doi = {10.1007/978-3-319-42432-3_37},
}

@InProceedings{Aric_al_netx,
  author     =  {Aric A. Hagberg and Daniel A. Schult and Pieter J. Swart},
  title      =  {Exploring Network Structure, Dynamics, and Function using NetworkX},
  booktitle  =  {Proceedings of the 7th Python in Science Conference},
  pages      =  {11 - 15},
  address    =  {Pasadena, CA USA},
  year       =  {2008},
  editor     =  {Ga\"el Varoquaux and Travis Vaught and Jarrod Millman},
  note       = {Version. 3.4.2}
}

@article{sullivan2019pyvista,
  doi = {10.21105/joss.01450},
  url = {https://doi.org/10.21105/joss.01450},
  year = {2019},
  month = {May},
  publisher = {The Open Journal},
  volume = {4},
  number = {37},
  pages = {1450},
  author = {Bane Sullivan and Alexander Kaszynski},
  title = {{PyVista}: {3D} plotting and mesh analysis through a streamlined interface for the {Visualization Toolkit} ({VTK})},
  journal = {Journal of Open Source Software},
  note    = {Version. 0.45.2}
}

@misc{python-fcl,
  author       = {Berkeley Automation},
  title        = {python-fcl: Python bindings for the Flexible Collision Library (FCL)},
  year         = {2023},
  howpublished = {\url{https://github.com/BerkeleyAutomation/python-fcl}},
  note         = {Version. 0.7.0.8}
}

@misc{trimesh,
  author    = {Dawson-Haggerty, Michael and others},
  title     = {Trimesh: A Python library for loading and using triangular meshes with an emphasis on watertight surfaces},
  year         = {2022},
  howpublished = {\url{https://trimesh.org/}},
  note      = {Version 4.6.13}
}

@misc{meshlib,
  author       = {MeshInspector},
  title        = {MeshLib: An SDK to Supercharge Your 3D Data Processing Efficiency},
  year         = {2025},
  howpublished = {\url{https://pypi.org/project/meshlib}},
  note         = {Version. 3.0.6.229}
}

@book{craig2005intro_rob,
  title={Introduction to Robotics: Mechanics and Control},
  author={Craig, John J.},
  edition={3rd},
  year={2005},
  publisher={Pearson Prentice Hall},
  address={Upper Saddle River, NJ},
  isbn={0-13-123629-6}
}

@article{yen1971kshortest,
  author  = {Yen, Jin Y.},
  title   = {Finding the K Shortest Loopless Paths in a Network},
  journal = {Management Science},
  volume  = {17},
  number  = {11},
  pages   = {712--716},
  year    = {1971},
  publisher = {INFORMS},
  url     = {https://www.jstor.org/stable/2629312}
}

\appendix

\section{Case study: DFM extraction}
\label{app:dof_case}

\begin{figure}[h]
\centering
\includegraphics[width=0.2\textwidth]{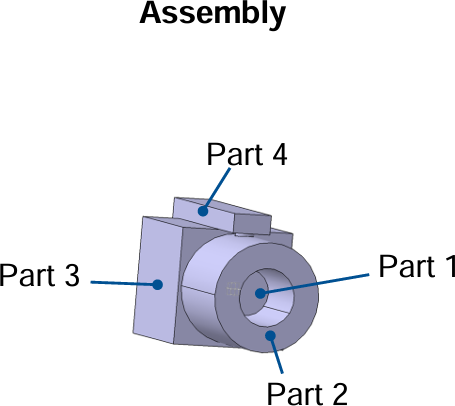}
\caption*{Figure A1: The assembly for the case study.}
\label{fig:assembly_case_study}
\end{figure}

To clearly illustrate the \ac{dof} extraction method described in Section \ref{sec:methods}, we constructed an assembly comprising four parts, as shown in Figure A1.
We used this assembly to show our approach in detail.
We first export these CAD models into STL format. 
Then, spatial relationship detection is applied to build the relational matrix. 
We denote the contact, blocking, and free relationships as 1, 2, and 3, respectively. 
The relationship between one part and itself is denoted as zero. 
The relational matrix $RM$ is as \eqref{eq:relational_matrix_case_study}.

\begin{equation}
\small
RM = \begin{pmatrix}
 0 & 2 & 1 & 1 \\
 2 & 0 & 1 & 1 \\
 1 & 1 & 0 & 1 \\
 1 & 1 & 1 & 0
\end{pmatrix}
\label{eq:relational_matrix_case_study}
\end{equation}

We define relevant joint coordinates for the next step and perform geometric constraint construction. 
For instance, the DoF matrix of $part \ 1$ with $joint \ 13$ under $coordinate \ system \ 1$ is as \eqref{eq:dof_matrix_case_study}.

\begin{equation}
\small
DoFM = \begin{pmatrix}
 1 & 1 & 0 & 0 \\
 1 & 1 & 0 & 0 \\
 1 & 0 & 1 & 1
\end{pmatrix}
\label{eq:dof_matrix_case_study}
\end{equation}
\section{Assembly planning directed graph algorithm}
\label{app:algo}
This appendix outlines the algorithm used to generate the weighted directed graph $D(V, E, W)$, given a part graph $G(N, J)$.
The algorithm iteratively explores the solution space by simulating the disassembly process (removing joints from the part graph) to generate layers in the directed graph.
It strictly enforces the Single-Piece Flow constraint and utilizes \ac{dof} matrices to check for collisions during joining and prune geometrically infeasible sequences.

In principle, the framework handles parallel assemblies by removing the Single-Piece Flow check (Line 9 of Algorithm 1).
We generated the weighted assembly directed graph for Assembly 1, without the Single-Piece Flow heuristic.
The graph including parallel assemblies consisted of 8,192 nodes and 53,248 edges, consistent with the number of edges calculated from Eq.~\ref{eq:max-edges}.

\begin{algorithm}
\footnotesize
\caption{Assembly weighted directed graph generation}\label{alg:assemly-digraph-generation}
\begin{algorithmic}[1]

\Require Assembly Graph $G_{part} = (N, J)$, $DoF$, Weights $\mu$
\Ensure Weighted Directed Graph $D = (V, E, W)$

\State $V_{layers} \gets \{\}$ \Comment{Stores valid cutsets per layer}
\vspace{-0.0\baselineskip} 
\State $D \gets \text{empty directed graph}$ 
\vspace{-0.0\baselineskip} 
\State $L \gets |J|$ \Comment{Total number of layers}
\vspace{-0.0\baselineskip} 

\State $G_{sub} \gets G_{part}$
\vspace{-0.0\baselineskip} 
\For {$k \gets 0 : L-1$} \Comment{Iterate through layers}
\vspace{-0.0\baselineskip} 
\State $V_{layers}[k+1] \gets []$
\vspace{-0.0\baselineskip} 

\ForAll{subsets $J_{rem} \subseteq J$ where $|J_{rem}| = k+1$}
\vspace{-0.0\baselineskip} 
\State $G_{sub} \gets \text{remove\_edges}(G_{part}, J_{rem})$
\vspace{-0.0\baselineskip}

\If{\text{IsConnected}(G\_{sub})} \Comment{Single-Piece Flow Constraint}
\vspace{-0.0\baselineskip}
\State $V_{layers}[k+1] \gets V_{layers}[k+1] \cup \{G_{sub}.edges()\}$
\vspace{-0.0\baselineskip}

\ForAll {$E_{prev} \in V_{layers}[k]$}
\vspace{-0.0\baselineskip} 

\If{is\_subset($G_{sub}.edges, E_{prev}$)}
\vspace{-0.0\baselineskip} 
\If{\text{CollisionFree}($G_{sub}, \text{DoF}$)} \Comment{DoF check}
\vspace{-0.0\baselineskip} 

\State $e \gets \text{create\_transition}(E_{prev}, G_{sub})$ 
\State $w_e \gets \text{Eq.}~\eqref{eq:edge_weight}$ \Comment{Calculate $\mu$-weighted cost}
\State $D.\text{add\_edge}(e, w_e)$
\vspace{-0.0\baselineskip} 

\EndIf
\vspace{-0.0\baselineskip} 
\EndIf
\vspace{-0.0\baselineskip} 
\EndFor
\vspace{-0.0\baselineskip} 
\EndIf
\vspace{-0.0\baselineskip} 

\State $G_{sub} \gets \text{restore\_edges}(G_{part}, J_{rem})$

\EndFor
\vspace{-0.0\baselineskip} 

\State $\text{prune\_dead\_ends}(V_{layers}[k+1])$
\Comment{Remove nodes without successors}

\EndFor

\State \Return $D$

\end{algorithmic}
\end{algorithm}

\section{Dataset specification for Assembly 2}
\label{app:ass_2}
This appendix presents Assembly 2, comprising 17 joints (Fig. C2) and 15 parts (Fig. C3). 
It serves as the high-complexity benchmark for the scalability experiment presented in Subsection \ref{subsec:comp-eff}.

\begin{figure}[h]
    \centering
    \label{fig:ass_2_labels}
    \includegraphics[width=0.8\linewidth]{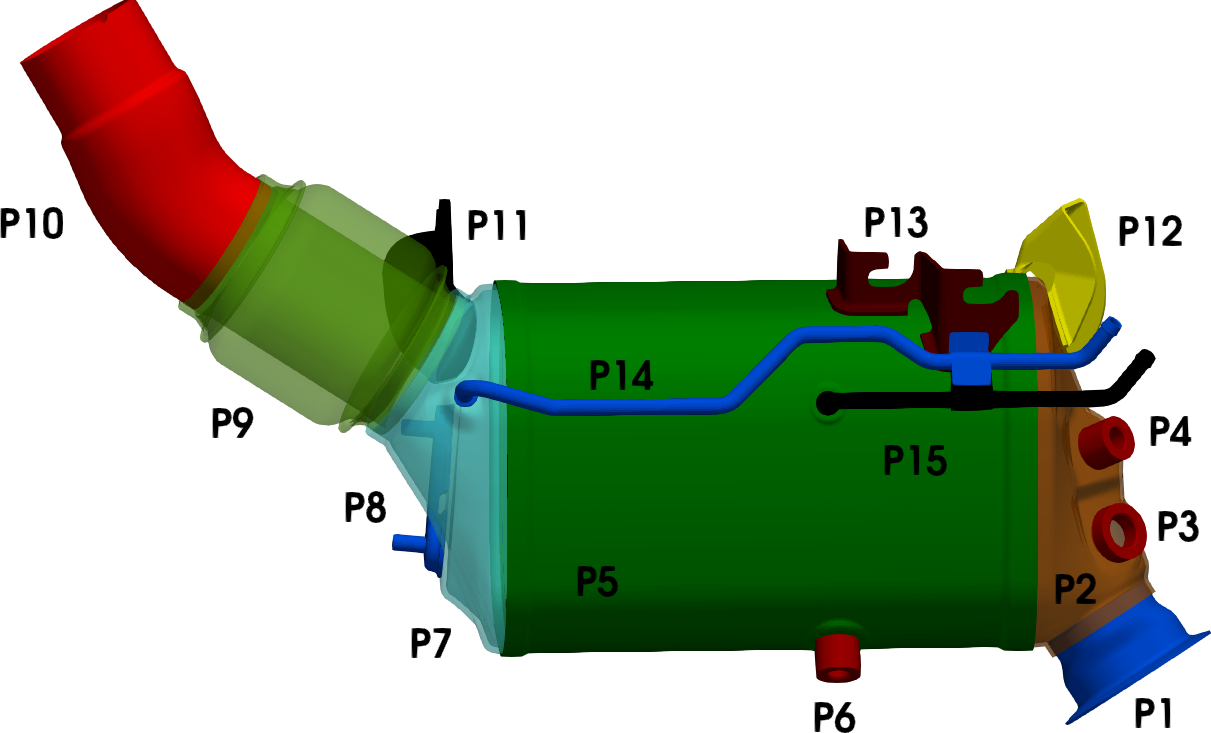}
    \caption*{Figure C1: Assembly 2 geometry and part labels.}
\end{figure}

\begin{figure}[H]
    \label{fig:ass_2_edges}
    \includegraphics[width=0.48\textwidth]{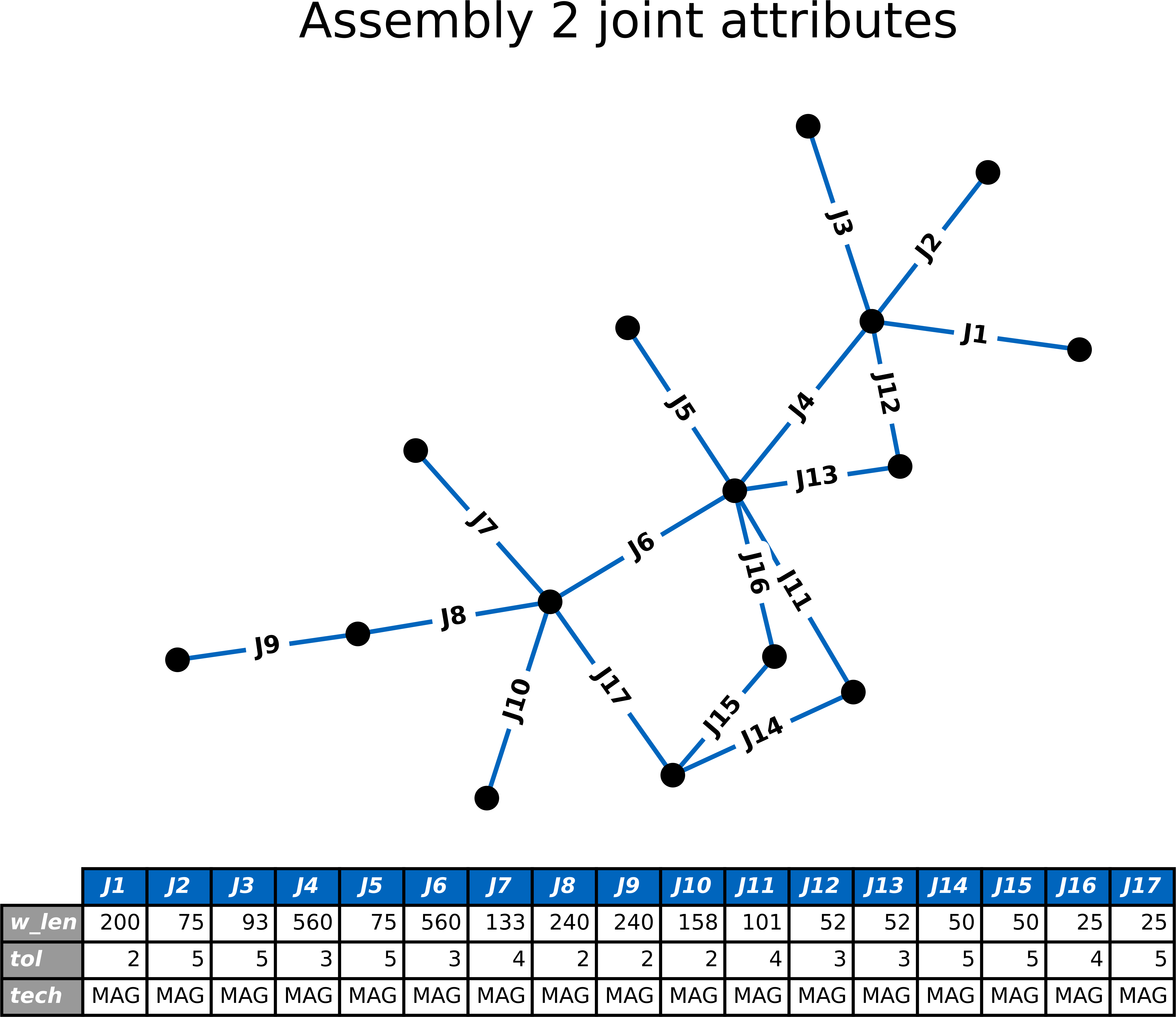}
    \caption*{Figure C2: Graph representation of the connections (edges) for Assembly 2. Edge labels indicate the welding length [mm] (used as operation time in our model), tolerance value (integer 1-5, where higher values indicate stricter tolerances), and the joining technology type (only "MAG").}
\end{figure}

\begin{figure}[H]
    \label{fig:ass_2_nodes}
    \includegraphics[width=0.48\textwidth]{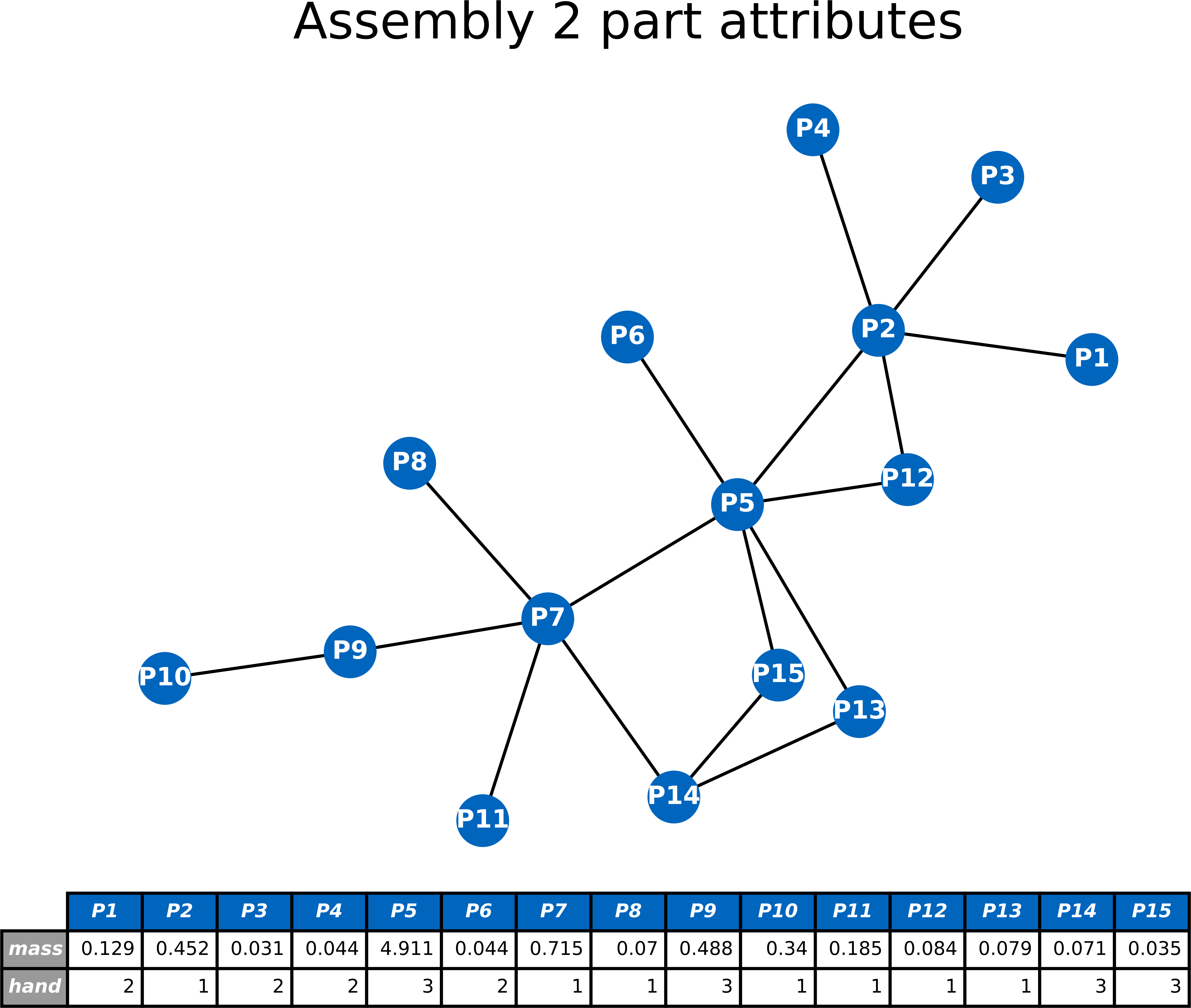}
    \caption*{Figure C3: Graph representation of the components (nodes) for Assembly 2. Node labels indicate component mass [kg] and the handling complexity level (integer 1-3, where higher values indicate higher fragility).}
\end{figure}

\end{document}